%% file: main.tex
\documentclass{article}
\usepackage{fullpage}

\PassOptionsToPackage{numbers,round}{natbib}
\usepackage[authoryear]{natbib}

\usepackage[utf8]{inputenc}
\usepackage[T1]{fontenc}
\usepackage{hyperref}
\usepackage{url}
\usepackage{booktabs}
\usepackage{amsfonts}
\usepackage{nicefrac}
\usepackage{microtype}
\usepackage{xcolor}
\usepackage{pgfplots}
\usepackage{tabstackengine}
\DeclareUnicodeCharacter{2212}{−}
\usepgfplotslibrary{groupplots,dateplot}
\usetikzlibrary{patterns,shapes.arrows}
\pgfplotsset{compat=newest}

\usepackage{mathtools}
\usepackage{amsmath, amssymb, bm}
\usepackage{amsthm}
\usepackage{subcaption}

\usepackage{enumitem}
\setlist[itemize]{align=parleft,left=0pt..1em}

\newtheorem{theorem}{Theorem}

\newtheorem{lemma}[theorem]{Lemma}

\theoremstyle{definition}

\newtheorem{remark}{Remark}
\newtheorem{assumption}{Assumption}

\newcommand{\ba}{\bm{a}}
\newcommand{\bb}{\bm{b}}

\newcommand{\bbm}{\bm{m}}

\newcommand{\bu}{\bm{u}}
\newcommand{\bv}{\bm{v}}
\newcommand{\bw}{\bm{w}}
\newcommand{\bx}{\bm{x}}
\newcommand{\by}{\bm{y}}

\newcommand{\cD}{\mathcal{D}}

\newcommand{\bbP}{\mathbb{P}}
\newcommand{\bbQ}{\mathbb{Q}}

\newcommand{\mtp}{\mathrm{MTP}}

\newcommand{\one}{\boldsymbol{1}}

\newcommand{\psd}{\mathbb{S}_+}

\newcommand{\bEx}{\ensuremath{\mathbb{E}}}

\newcommand{\ex}[1]{\ensuremath{\mathbb{E}\left[ #1\right]}}

\DeclareMathOperator{\diag}{\mathrm diag}

\DeclareMathOperator{\gtr}{tr}

\DeclareMathOperator{\gvec}{\mathsf{vec}}

\DeclareMathOperator{\mse}{\mathsf{mse}}

\let\originalleft\left
\let\originalright\right
\renewcommand{\left}{\mathopen{}\mathclose\bgroup\originalleft}
\renewcommand{\right}{\aftergroup\egroup\originalright}

\newcommand{\reals}{\mathbb{R}}

\newcommand{\eps}{\epsilon}
\newcommand{\normal}{\mathcal{N}}

\title{Fundamental limits for rank-one matrix estimation \\ with groupwise heteroskedasticity}

\author{Joshua K. Behne \thanks{Department of Electrical and Computer Engineering, Duke University} \and Galen Reeves \thanks{Department of Electrical and Computer Engineering and the Department of Statistical Science, Duke University}}

\begin{document}

\maketitle

\begin{abstract}

Low-rank matrix recovery problems involving high-dimensional and heterogeneous data appear in applications throughout statistics and machine learning. The contribution of this paper is to establish the fundamental limits of recovery for a broad class of these problems. In particular, we study the problem of estimating a rank-one matrix from Gaussian observations where different blocks of the matrix are observed under different noise levels. In the setting where the number of blocks is fixed while the number of variables tends to infinity, we prove asymptotically exact formulas for the minimum mean-squared error in estimating both the matrix and underlying factors. These results are based on a novel reduction from the low-rank matrix tensor product model (with homogeneous noise) to a rank-one model with heteroskedastic noise.

As an application of our main result, we show that recently proposed methods based on applying  principal component analysis (PCA) to weighted combinations of the data are optimal in some settings but sub-optimal in others. We also provide numerical results comparing our asymptotic formulas with the performance of methods based on weighted PCA, gradient descent, and approximate message passing.

\end{abstract}

\section{Introduction}

The problem of extracting information from noisy pairwise data has been studied extensively by focusing on spiked matrix models in which a  low-rank signal matrix (the spike) is observed in additive Gaussian noise. The theoretical properties of these models are now well understood in the setting where all of the interactions are of the same type and thus the noise has constant variance. However, much less is known about settings in which different types of information are obtained about different types of pairwise interactions.

\paragraph{Groupwise spiked matrix model}
This paper provides an asymptotically exact information-theoretic analysis for a general setting in which the underlying objects can be partitioned into to different groups and the strength of the pairwise interactions depends on these groups. Specifically, we focus on the $K$-group spiked matrix model given by
\begin{align}
    Y_{k\ell} = \sqrt{\frac{ \lambda_{k\ell}}{N}} \bx_k \bx_\ell^\top + W_{k\ell} , \quad k,\ell = 1, \dots  K \label{eq:Ykl}
\end{align}
where each $\bx_k$ is an unknown $n_k \times 1$ vector of latent random variables and each $W_{k\ell}$ is an unknown $n_{k} \times n_{\ell}$ noise matrix with independent standard Gaussian entries. The signal-to-noise ratio in each block is parameterized by a non-negative number $\lambda_{k\ell}$ and a global scale parameter $N$, which are assumed to be known. Given the collection of observations $Y = (Y_{k\ell})$  and the $K \times K$ matrix $\Lambda = (\lambda_{k\ell})$, the goal is to estimate the latent vectors, $(\bx_k)$, as well as their rank-one outer products, $(\bx_k \bx_\ell^\top)$. 

This model can be expressed in matrix form as
\begin{align}
\label{eq:Y}
    Y =  G \circ \bx \bx^\top
    + W,
\end{align}
where $G = (G_{k\ell})$ is a block constant matrix whose $k\ell$-th block has value  $\sqrt{ \lambda_{k\ell}/N}$, 
$\circ$ denotes the Hadamard (entry-wise) matrix product, and $\bx$ is column vector of dimension $n_1 + \dots + n_K$ obtained by stacking $\bx_1, \dots, \bx_K$. We emphasize that $G$ is a known parameter of the model. Hence, if the $\lambda_{k\ell}$ are strictly positive, then each block of the observations can be rescaled to obtain an equivalent model in which the rank-one signal matrix $\bx \bx^\top$ is corrupted by independent additive Gaussian noise whose variance is constant over blocks.

\subsection{Applications}
\label{sec:applications}

Our theoretical results for the groupwise model in \eqref{eq:Ykl} have implications for a variety of applications. In the following, we sometimes use the notation $\bu$ and $\bv$ instead of $\bx_k$ to represent the latent vectors.

\paragraph{PCA with heteroskedastic data}
Generalizations of the spiked covariance matrix model of \cite{johnstone:2001a} have been used to study the effects of heteroskedastic noise \citep{bai:2012,yaonbai:2015,hong:2018pca,hong:2018,zhang:2021}. A rank-one version of such a model can be described as follows. Conditional on a $d \times 1$ latent vector $\bu$, observations are generated independently according to
\begin{align}
\label{eq:hetroPCA}
    \by_i = \sqrt{\frac{1}{N}}\, \bu v_i + \eta_i \bm{\epsilon}_i, \quad i = 1, \dots, n 
\end{align}
where each $v_i$ is a standard Gaussian variable, each $\bm{\eps}_i$ is a $d \times 1$ vector with independent standard Gaussian entries, and the terms $N$ and $\eta_1, \dots , \eta_L$ are deterministic parameters.

\cite{hong:2018} provide asymptotically exact formulas for estimating $\bu$ using weighted PCA in the high-dimensional setting where $d,n \to \infty$, while the noise levels $\eta_i$ are supported on fixed set of $L$ values $\{\sigma_1, \dots, \sigma_L\}$. Furthermore, in the setting where the noise levels are known, they describe the optimal choice of weights and the corresponding error rates. Related work by \cite{zhang:2021} focuses on a version of \eqref{eq:hetroPCA} in which the role of $\bu$ and $\bv = (v_1, \dots, v_n)$ are interchanged and the noise levels are unconstrained.

The model \eqref{eq:hetroPCA}  can be seen as a special case of the groupwise model in \eqref{eq:Ykl} with $K=L + 1$ where the observations are grouped according to common noise levels such that
\begin{align}
\label{eq:HPCA_group}
    Y_{\ell} = \sqrt{\frac{1}{N}} \bu \bv^\top_{\ell} + \sigma_\ell W_{\ell}, \qquad \ell = 1, \dots, L
\end{align}
where $\bv_\ell$ is an $n_\ell \times 1$ vector with independent standard Gaussian entries, $W_\ell$ is an $d \times n_{\ell}$  matrix with independent standard Gaussian entries,  and $n_{\ell} \coloneqq | \{ i  : \eta_i =\sigma_\ell\}|$. Here, we have used a slightly different convention for indexing the groups because each observation involves the same factor $\bu$ (i.e., the interactions between the $\bv_\ell$ terms are not observed).

As a direct consequence of our main result, we obtain asymptotically exact formulas for the fundamental limits of estimating $\bu$ in the setting where the noise levels are known. In the case where $\bu$ has IID Gaussian entries, we show that the fundamental limits match the achievable results obtained by \cite{hong:2018}, thus demonstrating that weighted PCA is information-theoretically optimal in this setting; see Theorem~\ref{thm:wpca}.

\paragraph{Covariate assisted clustering}
Another application of interest involves  clustering problems with high-dimensional covariates \citep{binkiewicz:2017,deshpande:2018,arroyo:2020}. As a specific example, consider the Gaussian version of the contextual stochastic block model introduced by \cite{deshpande:2018}, where an unknown random vector $\bu \in \{\pm 1\}^{d}$ has independent Radamacher variables (corresponding to the assignment of each entry to one of two equiprobable communities) and the observations consist of a symmetric $d \times d$ matrix  of pairwise observations and an asymmetric $d \times n$ matrix of interactions with an unknown $n \times 1$ latent vector $\bv$ 
\begin{subequations}
\label{eq:csbm}
\begin{align}
Y_{uu} &= \sqrt{\frac{\lambda_{uu}}{N}} \bu \bu^\top + W_{uu} \\
Y_{uv} &= \sqrt{\frac{\lambda_{uv}}{N}} \bu \bv^\top + W_{uv}.
\end{align}
\end{subequations}
\citet{deshpande:2018} characterize the statistical thresholds for detection and weak recovery of $\bu$ and show that these thresholds can be achieved using spectral methods. Ensuing work by \citet{ma:2021} proves exact formulas for the asymptotic MMSE and also provides a sharp asymptotic analysis of the MSE associated with an AMP algorithm adapted to this setting.

In view of our groupwise model \eqref{eq:Ykl}, we see that \eqref{eq:csbm} corresponds to the case of $K=2$ groups where the pairwise interactions between the entries in $\bv$ are not observed (i.e., $\lambda_{vv}=0$). Applying our main result to this setting recovers the fundamental limits obtained previously as a special case. More generally, our results also apply in setting where $\lambda_{vv}$ is nonzero and thus one can perform clustering jointly on the entries of $\bu$ and $\bv$; see Section~\ref{sec:methods}.

\paragraph{Computational-to-statistical gaps} The analysis of fundamental limits also plays an important role in understanding the interplay between estimation and computational complexity. In this paper we identify two settings where efficient methods are optimal. Theorem~\ref{thm:wpca} gives conditions for the optimality of weighted principal component analysis (WPCA) and our numerical results demonstrate a close correspondence between our asymptotic formulas and the empirical performance of various methods.

Beyond these settings, our result can also be used to study settings where there exists a gap between the optimal performance and efficient methods. Indeed, for a large family of related estimation problems, it is well known that the same formulas describing the optimal limits provide information about the behavior of approximate message passing \citep{zdeborova:2016b}. We leave this as a direction for future research.

\subsection{Comparison with Prior Work}

Low-rank matrix estimation problems have been studied extensively with applications in covariance estimation \cite{johnstone:2001a,johnstone:2009}, sparse principal component analysis (PCA) \cite{zou:2006}, clustering \cite{lesieur:2016,banks:2018}, and community detection  \cite{decelle:2011a,abbe:2018}. The bulk of the theoretical work has focused on one of two basic models: the \emph{spiked Wigner model}, which corresponds to a single diagonal block of the form $Y_{kk}$ in the groupwise model\footnote{The spiked Wigner model assumes a symmetric noise matrix with variance 2 on the diagonal and 1 on the off-diagonal.  This corresponds to an independent noise matrix with a factor two difference in  SNR.}, and the \emph{spiked Wishart model}, which corresponds to a single off-diagonal block of the form $Y_{kl}$ in the groupwise model. 

The statistical limits for these models were studied by \citet{lesieur:2015, lesieur:2017} who used the replica method to derive conjectured formulas for the asymptotic mutual information and MMSE in the setting of independent and identically distributed variables. These formulas were  proven rigorously  in ensuing work  \cite{krzakala:2016, barbier:2016a, lelarge:2018, miolane:2017a} using a variety of interpolation methods from statistical physics. A complementary line of work has focused on the algorithmic limits of recovery associated with spectral methods \cite{baik:2005,benaych-georges:2012, deshpande:2014b,perry:2018}, semidefinite relaxations \cite{amini:2009},  and approximate message passing (AMP) \cite{ fletcher:2018,deshpande:2014,parker:2014a,lesieur:2017,montanari:2021}. 

More recent work has focused on extending these models in a number of directions, including models with structured priors described by generative models  \cite{aubin:2021,cocola:2020,pandit:2020}, multiview models involving multiple different observations on the same underlying variables~\cite{mayya:2019, barbier:2020a, reeves:2020MTP}, and low-rank tensor observation problems \cite{luneau:2020,chen:2021}. Within this body of work, there has also been significant interest on the existence of computational-to-statistical gaps under various detection and recovery criteria \cite{barbier:2020,weed:2020}.

In a slightly different context, a special case of the groupwise model (involving noise levels with certain symmetries) has appeared previously as a proof technique  \cite{barbier2018rankone} where the goal was to establish an asymptotic equivalence between the standard spiked Wigner model and a \emph{spatially coupled} collection of rank-one models, whose fundamental performance could be analyzed rigorously via the analysis of AMP. 

Our theoretical results build upon a long line of work originally in the statistical physics literature including the adaptive interpolation method \citep{barbier:2018} and its recent generalizations of the usual low-rank matrix estimation problems \citep{barbier:2020a, reeves:2020MTP}. A key step in our analysis is a novel reduction between these models that does not seem to have been recognized previously.

\section{Fundamental Limits}

Our main results are  exact formulas for the asymptotic mutual information and MMSE associated with the groupwise model \eqref{eq:Ykl}. In addition to the pairwise observations, we also allow for side information of the form 
\begin{align}
    Y_k = \sqrt{r_k} \bx_k  +W_k, \qquad  k = 1, \dots, K \label{eq:Yk}
\end{align}
where each $W_k$ is an unknown $n_k \times 1$ noise vector with independent standard Gaussian entries. The entire collection of observations is denoted by $Y = (Y_k, Y_{k\ell})$.

\begin{assumption}\label{assumption:1} We consider a sequence of problems, indexed by an integer $N$, with observations  from the groupwise spiked model  \eqref{eq:Ykl} and the side information model  \eqref{eq:Yk}  such that:
\begin{enumerate}[align=parleft,left=.5em..2em,label=\emph{\Alph*})]
\item The number of groups $K$ is fixed while number of variables in each group $n_k = n_k(N)$ scales as $n_k/N \to \beta_k \in (0, \infty)$.

\item For each $k = 1, \dots ,K$, the entries of  $\bx_k = (x_{k1}, \dots, x_{kn_k})$ are drawn independently from a distribution $P_{k}$ on $\reals$ with second moment equal to one and finite fourth moment. 
\end{enumerate}
\end{assumption}

Rather than working with mutual information directly, we use an equivalent parameterization in terms of the relative entropy between the distribution of the observations and the standard Gaussian measure of the same dimension. For each group $k$, we define the single-letter relative entropy function $D_k \colon [0, \infty) \to [0, \infty)$ according to 
\begin{align} \label{eq:slre}
D_k(\gamma) \coloneqq D(\bbP_{\sqrt{\gamma}\, x_{k} + w_{k} } \, \| \, \bbP_{w_{k}} )
\end{align}
where $x_{k} \sim P_k$ and $w_{k} \sim \normal(0,1)$ are independent (univariate) random variables and  $D(\bbP \, \| \,  \bbQ) = \int \log (\frac{d \bbP}{ d \bbQ} ) \, d \bbP$ is the relative entropy (or Kullback-Leibler divergence) between probability measures $\bbP$ and $\bbQ$.
Associated with the entire collection of observations, we define  $\cD_N \colon[0, \infty)^K \times  [0, \infty)^{K \times K} \to [0, \infty)$  to be
\begin{align} \label{eq:cD_N}
    \cD_N(r, \Lambda) & \coloneqq \frac{1}{N} D\left( \bbP_Y \, \| \, \bbP_W \right).
\end{align}

Our first result shows that the relative entropy $\cD_N$ converges to a well-defined limit that depends only on the parameters $(\beta_k)$ and the single-letter entropy functions associated with each group.  For  vectors $u, v \in \reals^n$, we write $u \leq v$ if $u_i \leq v_i$ for $i = 1,\dots,n$.

\begin{theorem}\label{thm:entropy} 
Under Assumption~\ref{assumption:1}, the relative entropy function $\cD_N$ converges pointwise to the limit $\cD$ given by 
\begin{align} \label{eq:cD}
 \cD(r, \Lambda) \coloneqq \adjustlimits \max_{q } \inf_{\tilde{r}}  \bigg\{ \sum_{k=1}^K \beta_k D_{k}(r_k +\tilde{r}_k)  + \frac{1}{2}  q^\top \Lambda q
    - \frac{1}{2}  \tilde{r}^\top q  \bigg\}
\end{align}
where the maximum is over all $K$-dimensional vectors $q$ such that $0 \leq q \leq \beta$ and the infimum is over all $K$-dimensional non-negative vectors $\tilde{r}$.
\end{theorem}

The variational formula for the limit provides an explicit link between the prior information about the variables, encapsulated by the single-letter relative entropy functions, and the strengths of the different types of interactions, described by the matrix $\Lambda$. 

From the point of view of estimation, the significance of Theorem~\ref{thm:entropy} is that changes in the relative entropy with respect to $(r, \Lambda)$ can be related to the MMSE in estimating the underlying variables. Specifically, as a consequence of the I-MMSE relation \citep{guo:2005a}, one finds that $\cD_N$ is differentiable on the interior of its domain and its partial derivatives satisfy 
\begin{align} \label{eq:IMMSE}
\begin{split}
& \partial_{r_k} D_N(r, \Lambda) = \frac{1}{2N}  \ex{ \| \ex{ \bx_k \mid Y} \|_2^2} \\ & \partial_{\lambda_{k\ell}} D_N(r, \Lambda) = \frac{1}{2 N^2}  \ex{ \| \ex{ \bx_k\bx_\ell^\top \mid Y} \|_F^2}.
\end{split}
\end{align}
In conjunction with Theorem~\ref{thm:entropy}, these relations lead to exact formulas for the asymptotic MMSE. The main difference in the asymptotic setting is that  there may be a countable number of points at which the limit is non-differentiable. These points correspond to jump-discontinuities in the asymptotic MMSE.

\begin{theorem}\label{thm:mmse}
Consider Assumption~\ref{assumption:1} and suppose that for a given pair $(r,\Lambda)$  the maximum in  \eqref{eq:cD} is attained at a unique point $q^*$. For each $k$ such that $q^*_k =0$ or $r_k > 0$ the MMSE  satisfies
\begin{align*}
\lim_{N \to \infty} \frac{\bEx \| \bx_k - \ex{ \bx_k | Y} \|_2^2}{n_k} = 1  - \frac{q^*_k}{\beta_k},
\end{align*}
and for each pair $(k, \ell)$ such that $q^*_k q^*_\ell = 0$ or  $\lambda_{k\ell} + \lambda_{\ell k} > 0$, the MMSE of $\bx_k \bx_\ell^\top$ satisfies 
\begin{align*}
 \lim_{N \to \infty} \frac{\bEx \| \bx_k\bx_\ell^\top  - \ex{ \bx_k\bx_\ell^\top  | Y} \|_F^2}{n_k n_\ell } = 1  -\left( \frac{q^*_k} {\beta_k } \right)\! \left( \frac{q^*_\ell} {\beta_\ell } \right).
\end{align*}
\end{theorem}

We emphasize that whether the maximum in \eqref{eq:cD} is attained at unique point is a property that can be verified on a case by case basis. Furthermore, in Appendix~\ref{sec:uniqueness} it is shown that for any fixed $r$ (possibly zero) and fixed collection of off-diagonal terms $(\lambda_{k\ell} : k \ne \ell)$ the set of diagonal terms $(\lambda_{kk} : k =1, \dots, K)$ such that the mazimizer is not unique has Lebesgue measure zero. More generally, there may be cases where some of the limits holds even though the maximizer is not unique. 

The reason we need further assumptions on $(q^*, r, \Lambda)$ is related to the fact that a phase transition with respect to arbitrarily small changes in $(r, \Lambda)$ occurs when the global maximizer jumps from one location to another. For a more thorough discussion of this behavior see \cite[Section~II-B]{reeves:2020MTP}. While the conditions we provide are sufficient, they are probably not necessary.

Finally, there may be settings of interest where $r$ is zero and $\Lambda$ is sparse for which the assumptions of Theorem~\ref{thm:mmse} are not satisfied. The next result shows that every global maximizer provides a lower bound on the MMSE. This result is used in Section~\ref{sec:WPCA} to establish the optimally of weighted PCA.

\begin{theorem} \label{thm:bounds}
Consider Assumption~\ref{assumption:1} and suppose that for a given pair $(r, \Lambda)$ the maximum in \eqref{eq:cD} is attained at $q^*$. Then, the MMSE satisfies
\begin{align*}
\liminf_{N \to \infty} \frac{\bEx \| \bx_k - \bEx[ \bx_k \mid Y] \|_2^2 }{n_k}   &\ge  1  - \frac{q^*_k}{\beta_k},\\
\liminf_{N \to \infty} \frac{ \bEx \| \bx_k\bx_\ell^\top  - \bEx[ \bx_k\bx_\ell^\top  | Y] \|_F^2}{n_k n_\ell }  
&\ge 1  -\left( \frac{q^*_k} {\beta_k } \right) \!\left( \frac{q^*_\ell} {\beta_\ell } \right).
\end{align*}
\end{theorem}

To help interpret these results it is useful to consider some special cases. Observe that if there is no side information ($r=0$) and  $\Lambda$ is diagonal,  then the groupwise model decouples into $K$ independent spiked Wigner models. In this case, the limit  can be expressed as
\begin{align}
    \sum_{k=1}^K \max_{q_k \in [0,\beta_k]}  \left\{ \beta_k D_k( 2\lambda_{kk} q_k) - \frac{1}{2} \lambda_{kk} q_k^2 \right\}.
\end{align}
Here,  each summand corresponds to the limits obtained previously for the spiked Wigner model \cite{barbier:2016a} with the only  difference being a factor of 2 in the signal-to-noise ratio which arises because our model does not assume symmetric noise. 

Alternatively, if there is no side information, $\Lambda$ is anti-diagonal, and $K$ is even, then the groupwise model decouples into $K/2$ independent Wishart models. After some straightforward manipulations, the limit can be expressed as
\begin{align}
    \sum_{k < \ell \, : \, k + \ell = K} \adjustlimits \max_{q_k \in [0, \beta_k]} \min_{q_\ell \in [0, \beta_\ell]}  \left\{\beta_k D_{k}\left(2 \bar{\lambda}_{k\ell}  q_\ell \right ) + \beta_{\ell} D_{\ell}\left(2\bar{\lambda}_{k\ell} q_k \right )      - \bar{\lambda}_{k\ell} q_k q_\ell \right\}
\end{align}
where $\bar{\lambda}_{k\ell} = (\lambda_{k\ell} + \lambda_{\ell k})/2$. Here, each summand corresponds to the limit obtained previously for the spiked Wishart model \cite{miolane:2017a}.

\subsection{Proof Outline}\label{sec:proof_outline}
A key step in the proof of Theorem~\ref{thm:entropy} is to recognize that the groupwise model can be expressed as a special case of the matrix tensor product (MTP) model studied by \citet{reeves:2020MTP}. 
The MTP model was conceived as a generalization of multiview low-rank estimation models in which one obtains multiple observations associated with a  collection of $d$-dimensional variables represented by an $n \times d$ matrix $X$. The observations in this model have the form 
\begin{align}
\tilde{Y} =  (X \otimes X) B + \tilde{W} 
\end{align}
where $X\otimes X$ is the $n^2 \times d^2$ matrix formed by the Kronecker matrix product, $B$ is a known $d^2 \times m$  coupling matrix, and $\tilde{W}$ is an $n^2 \times m$ noise matrix with independent Gaussian entries.  The main result in \cite{reeves:2020MTP} is a formula describing the exact limit of the relative entropy in the asymptotic regime where the dimension $d$ is fixed while the number of variables $n$ increases to infinity. 

The connection between the groupwise model in \eqref{eq:Ykl} and the MTP  model can be understood as follows. First, let the latent vectors $\bx_1, \dots, \bx_K$ be embedded into an $n \times K$ block diagonal matrix of the form $X = \diag( \bx_1, \dots, \bx_K)$. Then, the matrix version of the groupwise model in \eqref{eq:Y} can be expressed as
\begin{align}
    Y = \frac{1}{\sqrt{N}} X \Gamma X^\top  + W, \qquad \Gamma \coloneqq (\sqrt{\lambda_{k\ell}}) .
\end{align}
Next we use vectorization to express this matrix equation in terms of the Kronecker product. For an $m \times n$ matrix $A$, let $\gvec(A)$ be the $mn \times 1$ vector obtained by stacking the columns of $A$. We have
\begin{align}
    \gvec(Y) = \frac{1}{\sqrt{N}} (X\otimes X) \gvec(\Gamma)  +\gvec(W). 
\end{align}
Hence, our model corresponds to the matrix tensor product where the dimension is equal to the number of groups and the  coupling ``matrix'' is described by the $k^2 \times 1$ vector $\gvec(\Gamma)$.

From here, the remaining  step in the proof is to simplify the expression for the limiting relative entropy. A direct application of the results in \citet{reeves:2020MTP} leads to a formula that requires optimization over the space of $K \times K$ symmetric positive semidefinite matrices (see \citep[Definition~3]{reeves:2020MTP}). Exploiting the block-diagonal structure of the matrix $X$ arising from the groupwise model, we are able to simplify this expression to obtain the formula given in Theorem~\ref{thm:entropy}, in which the optimization is taken over the space of $K$-dimensional nonnegative vectors. 

The proof of Theorem~\ref{thm:mmse} follows from Theorem~\ref{thm:entropy} using a similar approach as the proof of \cite[Theorem~2]{reeves:2020MTP}. The full proofs of Theorems~\ref{thm:entropy} and \ref{thm:mmse} are provided in Appendix~\ref{app:thment} and \ref{app:thmmmse} respectively.

\section{Optimality of Weighted PCA} \label{sec:WPCA}
This section considers some implications of our main results for the heteroskedastic model given in \eqref{eq:HPCA_group}. Recall that in this setting the observation consists of interactions between a latent vector $\bu$ and a collection Gaussian vectors $(\bv_\ell)$, where the noise level in each interaction is parameterized by a known value $\sigma_\ell$. Throughout this section we further assume that $d,n_1, \dots, n_L$ scale to infinity with the global scale parameter $N$ such that $d/N \to \beta_0 \in (0, \infty)$ and $n_\ell / N \to \beta_\ell \in (0, \infty)$. 

A standard approach for estimating $\bu$ from the  observations $Y_{1} \dots, Y_{L}$ is to use sample weighted principle component analysis (WPCA), where the estimate $\hat{\bu}$ is a scaled version of the leading eigenvector of the $d \times d$ positive semidefinite matrix 
\begin{align} 
    \sum_{\ell} \omega_\ell Y_\ell Y_\ell^\top \label{eq:wpcaYl}
\end{align}

\citet{hong:2018} study the asymptotic performance of WPCA for a low-rank version of \eqref{eq:HPCA_group} where $\bu$ is replaced by a matrix with orthogonal columns. For the rank-one setting considered in this paper, their results provide formulas for the squared correlation between $\hat{\bu}$ and $\bu$ as a function of the parameters $(\sigma_\ell, \beta_\ell) $ as well as the weights $(\omega_\ell)$. Furthermore, they derive the optimal choice of weights as a function of the problem parameters and provide a simplified formula for the corresponding correlation. Adapted to the notation of this paper, their results show that
\begin{align} \label{eq:hetpcaasc}
  \frac{\langle \hat{\bu}, \bu \rangle^2}{\|\bu\|^2 \|\hat{\bu}\|^2} \stackrel{a.s.} \longrightarrow q_0^{\text{WPCA}}/\beta_0
\end{align}
where $\hat{\bu}$ is obtained via the optimal choice of weights and $q_0^{\text{WPCA}}$ is given by the largest real root of the function
\begin{align}
    R(x) \coloneqq 1 - \sum_{\ell = 1}^L{\frac{\beta_\ell} {\sigma_\ell^2} \frac{\beta_0 - x}{\sigma_\ell^2 + x}}
\end{align}

To compare results for WPCA with the formulas for the MMSE obtained in this paper, observe that the model in \eqref{eq:HPCA_group} can be represented as an instance of the groupwise model \eqref{eq:Ykl} with $K = L+1$ groups indexed by the set $\{0,1, \dots, L\}$ via the mapping
\begin{align*}
\bu = \bx_0, \quad \bv_{\ell} = \bx_{\ell}, \quad 
    \lambda_{k\ell} = \begin{cases}  \sigma_{\ell}^{-2},  &  k = 0, \ell \ge 1\\
    0, & k \ge 1.
    \end{cases}
\end{align*}
Notice that in this case, the quadratic term in the variational formula \eqref{eq:cD} simplifies as
\begin{align}
    q^\top \Lambda q = q_0 \sum_{\ell = 1}^L  q_\ell /  \sigma_\ell^2   
\end{align}
Consequently, for $r = 0$ the expression \eqref{eq:cD} is given by 
\begin{align}
     \adjustlimits \max_{0 \leq q \leq \beta} \inf_{\tilde{r} \ge 0} & \left\{ \frac{1}{2} \sum_{k=0}^L \beta_k D_k(\tilde{r}_k) +  \frac{q_0}{2} \sum_{\ell = 1}^L{ \frac{q_{\ell}}{\sigma_\ell^2}}
    - \frac{1}{2} \tilde{r}^\top q \right\} \label{eq:opt_wpca}
\end{align}

\begin{theorem}
\label{thm:wpca}
Consider Assumption~\ref{assumption:1} and suppose that $\bu$ and $\bv_1, \dots, \bv_L$ have independent standard Gaussian entries. Then, \eqref{eq:opt_wpca} has a unique maximizer $q^*$. Furthermore, if 
\begin{align}
    \sum_{\ell = 1}^L{\frac{\beta_0 \beta_\ell}{\sigma_\ell^4}} > 1
\end{align}
then $q_0^* = q_0^\mathrm{WPCA}$. Otherwise $q_0^* = 0$.
\end{theorem}

Theorem~\ref{thm:wpca} shows that WPCA attains the fundamental limits described by our formulas. Moreover, it gives an explicit condition on the noise levels such that the correlation is strictly positive. The full proof of Theorem~\ref{thm:wpca} is given in Appendix~\ref{app:wpca}.

\section{Numerical Results}
\label{sec:methods}

To simplify the presentation of the algorithmic results, it is  convenient to apply a prepossessing step that symmetrizes the observations. The symmetrized observations and parameters are defined according to
\begin{align}
\begin{split}
& Y_{k\ell}^\mathrm{sym} = \sqrt{\frac{\lambda_{k\ell}}{\lambda_{k\ell} + \lambda_{\ell k}}} \,  Y_{k \ell} + \sqrt{\frac{\lambda_{\ell k}}{\lambda_{k\ell} + \lambda_{\ell k}}} \,  Y_{\ell k}^\top \\ & \lambda^\mathrm{sym}_{k \ell} = \lambda_{k\ell} + \lambda_{\ell k},
\end{split}
 \end{align}
where we use the convention  $Y_{k\ell}^\mathrm{sym}  = \sqrt{1/2} Y_{k\ell} + \sqrt{1/2} Y^\top_{\ell k}$ in the case $\lambda_{k\ell}^\mathrm{sym} =0$. 
By the orthogonal invariance of the standard Gaussian distribution, it can be verified that  $(Y^\mathrm{sym}_{k\ell})$ are sufficient statistics for estimation of $(\bx_k)$. The symmetric groupwise model can be written in matrix form according to 
\begin{align}
Y^\mathrm{sym} = G^\mathrm{sym} \circ \bx \bx^\top + W^\mathrm{sym},  \label{eq:Ysym}
\end{align}
where $\bx$ is the $n \times 1$ vector obtained from stacking the $(\bx_k)$,  $W^\mathrm{sym}$ is an $n \times n$ symmetric matrix drawn from the Gaussian orthogonal ensemble, and $G^\mathrm{sym}$ is an $n \times n$ block-constant matrix with entries in the $k\ell$-th block given by  $N^{-1/2} (\lambda^\mathrm{sym}_{k\ell})^{1/2}= N^{-1/2} (\lambda_{k\ell} + \lambda_{\ell k})^{1/2}$.

\subsection{Spectral Methods} \label{sec:spectral}
Spectral methods have been studied extensively both in the context of the spiked Wigner and spiked Wishart models, as well as covariate assisted models and multiview models. In the case of single-type interactions, the spectral methods are optimal when the factors have an orthogonally invariant distribution. We consider two baseline methods based on principal component analysis (PCA):
\begin{itemize}
    \item \textbf{Joint PCA:}
    Let $Y^*$ be the symmetric $n \times n$ data matrix obtained by zeroing out the blocks in $Y^\mathrm{sym}$ with $\lambda^\mathrm{sym}_{k\ell}=0$. The estimate of each $\bx_k$ is chosen to be proportional to the corresponding sub-vector in the  leading eigenvector of $Y^*$.
    \item \textbf{Weighted PCA:}  Given nonnegative weights $(w_{k\ell})$ the estimate of each $\bx_k$ is chosen to be proportional to the leading eigenvector of the $n_k \times n_k$ matrix
    \begin{align} \label{eq:kgroupwpca}
       w_{kk} Y_{kk}^\mathrm{sym}  + \sum_{ \ell \ne k} w_{k\ell} Y_{k\ell}^\mathrm{sym}  \left( Y_{k\ell}^\mathrm{sym} \right)^\top .
    \end{align}
    In the case of the two-group model, this method is equivalent to the one proposed by \citet[Theorem~6]{deshpande:2018}, who provide an explicit specification for the weights $w_{12}$ and $w_{21}$ as a function of the noise level in each block. Alternatively, if the $Y_{kk}$ observation is omitted (i.e.,  $w_{kk}=0$) then this reduces to the method studied in  
    Section~\ref{sec:WPCA}.
\end{itemize}

\subsection{Gradient Descent} \label{sec:gd}
Gradient descent algorithms have been studied in many contexts and provide a straightforward optimization procedure. Recent work (e.g., \cite{jzhang:2020,bhojanapalli:2016}) has shown that, under reasonable conditions, low-rank matrix recovery has an objective without any spurious local minima, even though it is a non-convex problem. A gradient method was also studied in the spiked matrix model under generative priors \cite{cocola:2020}.

We approximate the maximum likelihood estimate of the symmetric model \eqref{eq:Ysym} with gradient descent. Let $\bx^0$ be an initial estimate of $\bx$ and let $\gamma_0,\gamma_1, \dots$ be a sequence of positive numbers. For $t =0, 1,2,\dots$ the gradient descent updates are given by
\begin{align}
\bx^{t+1} = \bx^t  + \gamma_t \left[ 
\left(  G^\mathrm{sym} \circ Y^\mathrm{sym} \right)  \bx^t - \left( (G^\mathrm{sym})^{\circ 2}  (\bx^t)^{\circ 2} \right)  \circ \bx^t
\right]
\end{align}
where $(A)^{\circ2}$ denotes the entry-wise square for an arbitrary matrix or vector $A$. The updates can be expressed equivalently at the group level according to 
\begin{align}
\bx^{t+1}_k = \bx_k^t + \gamma_t \left[ \sum_{\ell =1}^K \sqrt{\frac{\lambda^\mathrm{sym}_{k\ell}}{N}} Y^\mathrm{sym}_{k\ell}  \bx^t_\ell  - \left( \frac{1}{N}  \sum_{\ell = 1 }^K     \lambda_{k \ell}^\mathrm{sym} \|\bx^t_\ell\|^2 \right) \circ \bx^t_k  \right]
\end{align}
where $k = 1, \dots, K$.

\subsection{Approximate Message Passing} \label{sec:AMP}

A variety of AMP algorithms have been proposed in the context of low-rank matrix estimation problems \cite{fletcher:2018,deshpande:2014,parker:2014a,lesieur:2017,montanari:2021,ma:2021}. This section gives a version of AMP adapted to the symmetric form of the groupwise model \eqref{eq:Ysym}. The derivation  follows the general outline given by \citet{lesieur:2017} and is described in full detail in Appendix~\ref{app:AMP}.

For $k =1, \dots, K$, define the function
\begin{align}
    \eta_k(a, b) = \frac{  \int x_k  \exp \left\{ a x_k - \frac{1}{2}b x_k^2 \right\}  dP_{k}(x_k) }{ \int \exp \left\{ ax'_k - \frac{1}{2} b (x'_k)^2 \right\} d P_{k}(x'_k)}, \label{eq:etak}
\end{align}
and let $\eta'(a,b)$ denote the derivative with respect to the first argument. Also, for each  $k=1,\dots, K$, let $\bbm_k^0$ be an initial estimate of $\bx_k$. For $t=0,1,2, \dots$ the AMP updates are given by
\begin{subequations}
\label{eq:ampkl}
\begin{align}
\ba^t_k &= \sum_{\ell =1}^K \sqrt{ \frac{\lambda_{k\ell}^\mathrm{sym}}{N}}  Y_{k\ell}^\mathrm{sym} \, \bbm^t_\ell - \left(\frac{1}{N}  \sum_{\ell =1}^K \lambda_{k\ell}^\mathrm{sym} (Y^\mathrm{sym}_{k\ell})^{\circ2}  \bv_\ell^t \right ) \circ \bbm_k^{t-1} \\
\bb_k^t &= \frac{1}{N} \sum_{\ell =1}^K\Big( \lambda^\mathrm{sym}_{k\ell}\,  (\bbm_\ell^t)^{\circ 2} + \lambda_{k\ell}^\mathrm{sym} (\one_{n_k \times n_{\ell}} - (Y^\mathrm{sym}_{k\ell})^{\circ 2})\,  \bv_\ell^t \Big) \\
    \bbm_k^{t+1}  &= \eta_k \left( \ba_k^t, \bb_k^t \right) \\
    \bv_k^{t+1} &= \eta_k' \left( \ba_k^t, \bb_k^t \right),
\end{align}
\end{subequations}
where $\bbm_k^{-1}$ and $\bv_k^0$ are initialized to the all zeros vector, the functions $\eta_k$ and $\eta'_k$ are applied entrywise to vector-valued inputs, and all of the terms for $k = 1,\dots,K$ are updated at each iteration. After $T$ iterations, the approximation of the conditional mean $\bx_k$ is given by $\bbm_k^T$ and the approximation to the conditional variance is given by $\bv_k^T$. Furthermore,  the vector $\ba_k^T$ can be viewed as an unbiased estimate of $\bx_k$. See Appendix~\ref{app:AMP} for more details.

\begin{figure*}[t]
\centering
    \begin{subfigure}{.48\textwidth}
      \input{figures/fig1a.tex}
      \caption{Gaussian prior}
      \label{fig:sfig1}
    \end{subfigure}
    \begin{subfigure}{.48\textwidth}
      \input{figures/fig1b.tex}
      \caption{Rademacher prior}
      \label{fig:sfig2}
    \end{subfigure}
  \caption{\label{fig:alpha_model} MSE in estimating the rank-one matrices $\bu\bu^\top$ and $\bv \bv^\top$ in the two-group model \eqref{eq:twogroup} as function of $\alpha$ with $n = 1024$ and $\lambda = 2$. The entries of $\bx$ are IID standard Gaussian in (a) and IID Rademacher in (b). The asymptotic MMSE (solid line) is given by Theorem~\ref{thm:mmse}. The diagonal block MSE is computed by averaging over 64 Monte Carlo trials for each algorithm.}
\end{figure*}
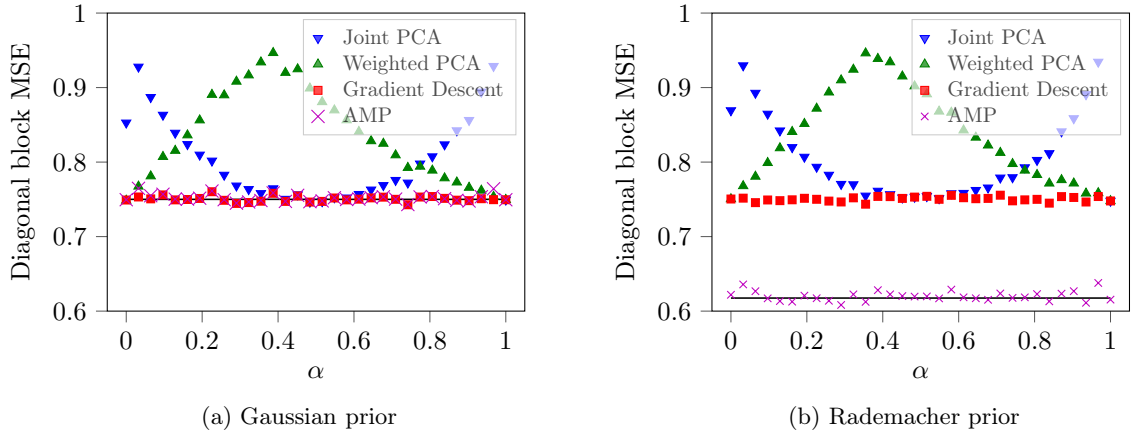

\subsection{Discussion of Empirical Results}

This subsection provides empirical results comparing the formulas for the asymptotic MMSE in Theorem~\ref{thm:mmse} and the empirical performance of the methods described in previous subsections. We will focus our attention on the groupwise model \eqref{eq:Ykl} with $K = 2$ groups and equal group sizes. We also parameterize our $\Lambda \in \reals^{2 \times 2}$ matrix by $\alpha \in [0, 1]$ and $\lambda \in (0, \infty)$. This yields the following model
\begin{align} \label{eq:twogroup}
    \Lambda(\alpha, \lambda) &= \lambda \begin{pmatrix} 1 - \alpha & \alpha \\ \alpha & 1 - \alpha \end{pmatrix}, \quad \bx = \begin{pmatrix} \bu \\ \bv \end{pmatrix}
\end{align}
where $\bu \in \reals^{n_u}$ and $\bv \in \reals^{n_v}$ with $n_u = n_v = n$. The MMSE for this model can be obtained from the global optimizer, $(q_1^*, q_2^*)$, of
\begin{align} \label{eq:twogprob}
 \adjustlimits \max_{q_1, q_2} \inf_{\tilde{r}_1 ,\tilde{r}_2} \left\{ \frac{1}{2}  D_{1}(\tilde{r}_1) + \frac{1}{2}  D_{2}(\tilde{r}_2) - \frac{1}{2}  (\tilde{r}_1 q_1 + \tilde{r}_2 q_2) + \frac{\lambda}{2} [  ( 1- \alpha)(q_1^2 + q_2^2)  + 2 \alpha q_1 q_2  ] \right\},
\end{align}
where $D_1$ and $D_2$ are the single-letter relative entropy functions associated with the distributions of $\bu$ and $\bv$, respectively, and we have taken the global scale parameter in Theorem~\ref{thm:entropy} to be $N = 2n$.

It can be observed that the fundamental limits of this model are invariant to the parameter $\alpha$ when the priors on $\bu$ and $\bv$ are the same. When this is the case in our empirical results, we will plot only a single curve for the MMSE (i.e., the curve generated by fixing any value of $\alpha$ and varying $\lambda$).

\paragraph{Evaluation of MSE}
Throughout our empirical results, performance is assessed in terms of the MSEs in estimating the  rank-one matrices $\bu \bu^\top$ and $\bv \bv^\top$ appearing on the diagonal blocks of $\bx \bx^\top$. In all cases, the estimates of the matrices are obtained from the estimates $\hat{\bu}$ and $\hat{\bv}$ of the corresponding vectors and the MSE can be expressed as
\begin{align}
\begin{split}
    & \mse_{uu} = \frac{1}{n^2} \ex{ \| \bu \bu^\top - \hat{\bu} \hat{\bu}^\top\|_F^2} \\ & \mse_{vv} =  \frac{1}{n^2} \ex{ \| \bv \bv^\top - \hat{\bv} \hat{\bv}^\top\|_F^2}
\end{split}
\end{align}
For the gradient descent method and AMP these expectations are approximated directly via Monte Carlo trials. For the spectral methods described in Section~\ref{sec:spectral}, the estimates of $\bu$ and $\bv$ are defined up to an unspecified scale factor. For the purposes of this paper, we focus on the best possible MSE that can be achieved under the constant-norm constraint $\|\hat{\bu}\| = \rho_u$ and $\|\hat{\bv}\| = \rho_v$ where $\rho_u$ and $\rho_v$ are fixed values that are optimized as a function of the distribution of the data. Starting with an initial estimate $\tilde{\bu}$ (assumed to be nonzero with probability one) the minimum MSE that can be obtained by projecting $\tilde{\bu}$ onto a sphere of fixed radius can be expressed as
\begin{align}
\inf_{\rho_u \ge 0 } \frac{1}{n^2}  \ex{ \left\| \bu \bu^\top -  \rho_u^2 \frac{\tilde{\bu} \tilde{\bu}^\top}{\|\tilde{\bu}\|^2} \right \|_F^2} = \frac{1}{n^2}\ex{  \|\bu\|^4}  - \frac{1}{n^2} \ex{ \frac{ ( \bu^\top \tilde{\bu})^2}{\|\tilde{\bu}\|^2}}^2.
\end{align}
Using this identity, we obtain an unbiased estimate of the desired MSE by approximating the second expectation on the right-hand side using Monte Carlo trials.

\paragraph{Implementation details}  The gradient descent method is initialized using estimates obtained from joint PCA, and AMP is initialized with each $\bbm_k^0$ initialized with IID $\normal(0,10^{-6})$ entries. For the weighted PCA method, the weights are chosen via a grid search optimization procedure that minimizes the empirical loss. Note that in doing so, the weights are adapted to the ground truth values of $\bu$ and $\bv$, and thus the resulting MSE should be viewed as a lower bound on the performance of the weighted PCA method for any data-driven selection of the weights. Additional details regarding the numerical approximation of the MMSE formulas are given in Appendix~\ref{app:MMSEApprox}. In addition, the code used to produce our numerical results can be found at \url{https://github.com/joshuabehne/GROME}.

\paragraph{Gaussian priors}
The case where both $\bu$ and $\bv$ have independent standard Gaussian entries is shown in Figure~\ref{fig:sfig1}. In this setting, the asymptotic MMSE is constant for all $0 \le \alpha \le 1$ and the weak detection threshold (i.e., the smallest value of $\lambda$ such that the asymptotic MMSE is strictly less than the  MMSE obtained without any observations) occurs at the critical threshold $\lambda^* =1$. The empirical performance of both gradient descent and AMP shows excellent agreement with the formulas for the asymptotic MMSE. By contrast, the spectral methods are sensitive to the choice of the  parameter $\alpha$. Additional plots for scenarios with Gaussian priors are given in Appendix~\ref{app:addfigs}.

\paragraph{Non-Gaussian priors}
A comparison of all algorithms considered in this paper is given in Figure~\ref{fig:sfig2} when both $\bu$ and $\bv$ have non-Gaussian priors. Specifically, $\bu$ and $\bv$ are both distributed IID according to a Rademacher distribution (i.e., each variable is $\pm 1$ with equal probability). We observe that only the AMP algorithm approaches the MMSE (which is invariant to $\alpha$ in this case) and the other methods (gradient descent, joint PCA, and weighted PCA) perform similarly to the Gaussian priors case, which can be seen in Figure~\ref{fig:sfig1}. Two additional examples involving non-Gaussian priors are shown in Appendix~\ref{app:addfigs}.

\begin{figure*}
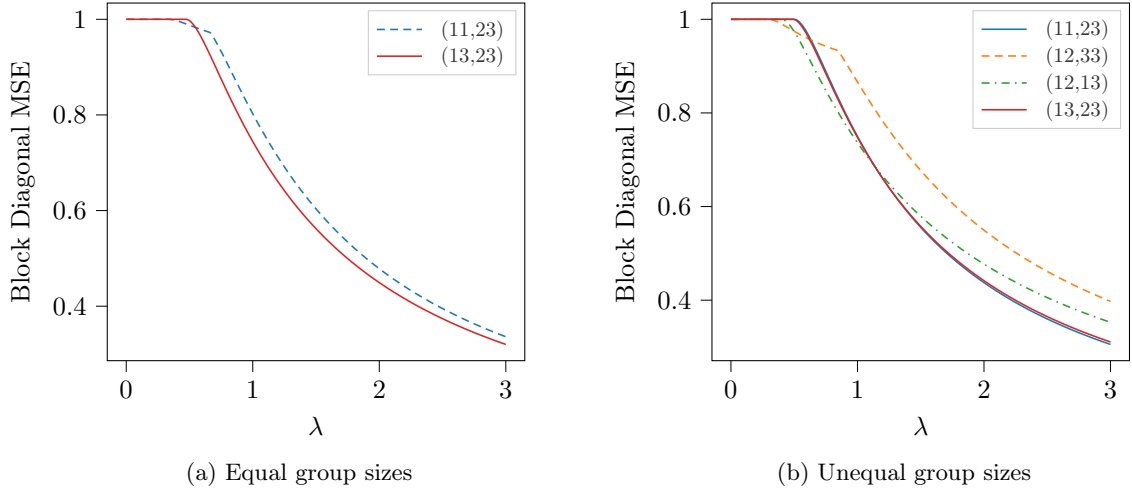

\centering
    \begin{subfigure}{.48\textwidth}
      \input{figures/fig4a.tex}
      \caption{Equal group sizes}
      \label{fig:threeequal}
    \end{subfigure}
    \begin{subfigure}{.48\textwidth}
      \input{figures/fig4b.tex}
      \caption{Unequal group sizes}
      \label{fig:threeunequal}
    \end{subfigure}
  \caption{\label{fig:threegroup} MMSEs for a three group model with Gaussian priors. The legend entry $(ij,k\ell)$ denotes the MMSE for the setup with signal parameters given by a $\Lambda$ matrix with indices of non-zero entries $ij$ and $k\ell$ and normalized to signal level $\lambda$, which is given by the x-axis values on both plots. Note that we only show a single example setup for each unique curve produced, i.e., certain combinations with support size two are omitted because they are redundant.}
\end{figure*}

\paragraph{Gaussian priors in a three group model}
The final experiment involves $K = 3$ groups with independent standard Gaussian priors. Both  equal groups sizes ($\beta_1 = \beta_2 = \beta_3 = 1/3$),  and unequal group sizes ($\beta_1 = 0.2,\beta_2 = \beta_3 = 0.4$) are considered. We restrict our attention to $\Lambda$ matrices such that 1) all the nonzero entries have the same value and 2) the total power satisfies $\beta^\top \Lambda \beta= \lambda$ for some  $\lambda > 0$. Under these assumptions, $\Lambda$ is specified uniquely by the pair $(\beta, \lambda)$ and the support (i.e., the indices of the nonzero entries in $\Lambda$).
For example, for the support $(11,23)$ we have 
\begin{align}
    \Lambda = \frac{\lambda}{\beta_1^2 + \beta_2 \beta_3} \begin{pmatrix} 1 & 0 & 0 \\ 0 & 0 & 1 \\ 0 & 0 & 0 \end{pmatrix}.
\end{align}

Figure~\ref{fig:threegroup} compares the asymptotic MMSE, as a function of $\lambda$, for all possible supports of size two. Notice that the weak recovery threshold no longer occurs at $\lambda = 1$. Interestingly, the different parameter settings are not ordered uniformly with respect to $\lambda$. In other words, the MMSE curves cross each other. One interpretation of these results is that focusing the signal power on interactions without any common factors improves performance at very low signal levels (in fact it lowers the weak recovery threshold) but degrades performance at higher signal levels.

The differences are more pronounced in the case of unequal group sizes  (Figure~\ref{fig:threeunequal}), where the effect of group $1$ being substantially smaller than the others dominates the behavior of some of the settings, namely those corresponding to $(11,23)$ and $(12,13)$.

\section{Conclusion}

This paper considers a block heterogeneous variation of the extensively studied low-rank matrix estimation problem. The information-theoretic limits for this model are derived by embedding it within the matrix tensor-product model studied by \citet{reeves:2020MTP}. We derive single-letter formulas for the MMSE that depend on the noise level in each block and the prior distributions of the underlying variables.  Complementary to the fundamental limits, we also consider the empirical performance of methods based on spectral decomposition, gradient decent, and AMP. 

The assumptions used in our theoretical results can be relaxed to a certain extent. For example, following the approach in \cite{reeves:2020MTP}, the number of groups $K$ can be allowed to grow at the rate $N^\alpha$ for sufficiently small positive number $\alpha$ and the assumption that the entries are identically distributed can also be relaxed provided that per-variable relative entropy function has a well-defined limit. 

An interesting open question for future work is whether a properly defined spectral method can achieve the weak recovery threshold in the two-group model \eqref{eq:twogroup} for all values of $\alpha$.

\section*{Acknowledgement}
This work was supported in part by NSF grant CCF-1750362.

\bibliographystyle{apalike}
\bibliography{AISTATS2022,long_names,library}

\clearpage

\appendix

\section{Proofs of Main Results} 
\label{sec:proofs}

As described in Section~\ref{sec:proof_outline} the main idea in the proof is to show that the groupwise model in \eqref{eq:Ykl} can be expressed as a special case of the matrix tensor product model studied in \cite{reeves:2020MTP}.  In the following, we first summarize the basic definition and results from \cite{reeves:2020MTP} and then provide the proofs for Theorems~\ref{thm:entropy} and \ref{thm:mmse}.

\subsection{The Matrix Tensor Product Model}
Let $X$ be an $n \times d$ random matrix. Following \citep[Equation~(9)]{reeves:2020MTP}, the matrix tensor product model associated with a positive scale factor $N$, $d \times d$, symmetric positive semidefinite matrix $R$,  and $d^2 \times d^2$ symmetric positive semidefinite matrix $S$ is defined by 
\begin{align}\label{eq:mtp}
    Y^\mtp_{R,S} & = \begin{dcases} 
       X R^{1/2} + W\\
    \frac{1}{\sqrt{N}} (X\otimes X) S^{1/2} + W'\\
    \end{dcases}
\end{align}
where $\otimes$ denotes the Kronecker matrix product and  $W$ and $W'$ are independent matrices with independent standard Gaussian entries. The only difference between this definition and the one in \cite{reeves:2020MTP} is that  the scale factor $N$ maybe different from the number of rows $n$.
The relative entropy function associated with this model is defined by 
\begin{align}
    \cD^\mtp_N(R,S) \coloneqq \frac{1}{N} D( \bbP_{Y^\mtp_{R,S}} \, \|\,  \bbP_{Y^\mtp_{0,0}}), \label{eq:cDmtp_N}
\end{align}
where $\bbP_{Y^\mtp_{0,0}}$ is the standard Gaussian measure of the same dimension as the observations. The approximation formula (see  \cite[Equation~(42)]{reeves:2020MTP}) is  defined as
\begin{align}
    \hat{\cD}^\mtp_N(R,S) \coloneqq \adjustlimits \max_{Q } \inf_{\tilde{R}  \in \psd^d} \left\{ \cD^\mtp_N(R +\tilde{R}, 0) +\frac{1}{2}  \gtr( S (Q\otimes Q)) - \frac{1}{2} \gtr(\tilde{R} Q)  \right\}. \label{eq:hatcDmtp_N} 
\end{align}
where the maximum is over $d \times d$ symmetric positive semidefinite matrices satisfying $Q \preceq \frac{1}{N} \ex{X^\top X}$. 

The main results  in \cite{reeves:2020MTP} are bounds on the difference between the relative entropy and the approximation formula.  For the purposes of this paper, we are interested in the asymptotic setting where the number of rows increases while the number of columns is fixed.

\begin{theorem}[{\cite{reeves:2020MTP}}]\label{thm:mtp}Consider a sequence of problems, indexed by integer $N$, with observations from the matrix tensor product model \eqref{eq:mtp} such that the number of columns $d$ is fixed while the number of rows scales as $n/N \to \beta \in (0,\infty)$. If the rows of $X$ are independent with  with $\ex{ |X_{ij}|^4} \le C$ for some constant $C$, then, $|\cD_N^\mtp - \hat{\cD}^\mtp_N|$ converges to zero pointwise as $N \to \infty$. 
\end{theorem}
\begin{proof}
This result follows along similar lines as  \cite[Corollary~1]{reeves:2020MTP}. If the entries of $X$ are bounded uniformly, then the result follows directly from \cite[Theorem~1]{reeves:2020MTP}. The relaxation from bounded entries to bounded forth moment follows from the continuity of relative entropy with respect to the quadratic Wasserstein distance \cite{polyanskiy:2016} and the fact that a finite fourth moment bound on the entries of $X$ implies a finite second moment bound on the entries of $X \otimes X$. 
\end{proof}

\subsection{Proof of Theorem~\ref{thm:entropy}} \label{app:thment}

Let $X = \diag( \bx_1, \dots ,\bx_K)$ be the $n \times K$ block-diagonal matrix whose $k$-th block is given by $\bx_k$ and let $\Gamma $ be the $K\times K$ matrix corresponding to the entrywise positive square root of $\Lambda$.  Using this notation, the groupwise model \eqref{eq:Ykl} can be expressed in matrix form according to
\begin{align}
 Y  =\frac{1}{\sqrt{N}} X \Gamma X^\top + W.
\end{align}
By vectorization, this model can be written equivalently as
\begin{align}
 \gvec(Y)  =\frac{1}{\sqrt{N}} (X\otimes X) \gvec(\Gamma) + \gvec(W).
\end{align}
This is the matrix tensor product model given in  \cite[Definition~1]{reeves:2020MTP}  with $K^2 \times 1$ coupling matrix $\gvec(\Gamma)$. As noted in \cite[Proposition~1]{reeves:2020MTP}, the fundamental limits for this model are the same as for the symmetric form of the matrix tensor product model (see \cite[Definition~2]{reeves:2020MTP}) characterized by $K^2 \times K^2$ coupling matrix $\gvec(\Gamma) \gvec(\Gamma)$. Recalling that the relative entropy function $\cD_N(r,\Lambda)$ in \eqref{eq:cD_N} is defined with respect to observations from the groupwise model \eqref{eq:Ykl} and the side-information model \eqref{eq:Yk} we can write
    \begin{align}
    \cD_N(r,\Lambda) =  \cD^\mtp_N(\diag(r),\gvec(\Gamma) \gvec(\Gamma)^\top ) \label{eq:cD_N_to_cDmpt_N}
    \end{align}
where $\cD_N^\mtp(R,S)$ is defined in \eqref{eq:cDmtp_N}. 

From here, a direct application of Theorem~\ref{thm:mtp} shows that the limiting behavior of $\cD_N$ can be expressed in terms of the approximation formula for the matrix tensor product in \eqref{eq:hatcDmtp_N}. To simplify the analysis, we use the following result, which leverages the block-diagonal structure in $X$.

\begin{lemma}\label{lem:diagS}
Suppose that $X = \diag(\bx_1, \dots, \bx_K)$ is a block-diagonal matrix where each $\bx_k$ is a column vector. Then, the relative entropy function in the matrix tensor product model depends only on the diagonal entries of the matrices $(R,S)$. Specifically, 
\begin{align}
\cD_N^\mtp(R,S) = \cD_N^\mtp(\diag(r) ,\diag(s)),
\end{align}
where $r$ and $s$ are vectors containing the diagonal entries in $R$ and $S$, respectively. 
\end{lemma}
\begin{proof} 
Consider the observation model $ A \bu  + \bw$ where $\bu$ is a $p \times 1$ random vector, $A = [a_{1}, \dots, a_K]$ is  an $m \times p$ deterministic matrix, and $\bw$ is an independent $m \times 1$ vector with standard Gaussian entries. Suppose that the entries of $\bu$ are deterministically zero except in the $k$-th location and let $Q_k$ be an $m \times m$ orthogonal matrix (chosen independently of $\bu$ and $\bw)$ such that the $k$-th row of $Q_k$ is given by $a_k/\|a_k\|$. Then, we can write
\begin{align}
    D(A \bu  +\bw  \, \| \, \bw) & = D( a_k u_k  + \bw \, \| \, \bw)\\
    & = D( Q_k a_k u_k  + Q_k \bw \, \| \, Q_k \bw)\\
    &
    = D( (\diag(\|a_1\|, \dots, \|a_K\|) \bu  + Q_k\bw \,  \| \, Q_k\bw )
\end{align}
where the second line follows from the invariance of relative entropy to one-to-one transformation and the third line follows from first noting that $Q_k a_k = \|a_k\|e_k$, where $e_k$ is the $k$-th standard basis vector, and then recalling that only the $k$th entry in $\bu$ is nonzero. By the orthogonal invariance of the standard Gaussian distribution $Q_k\bw$ is a standard Gaussian vector that is independent of $\bu$. Now suppose that $T$ is a $p \times p$ symmetric positive  semidefinite matrix with diagonal given by $t$. Setting $A = T^{1/2}$ we find that $(\|a_1\|^2, \dots, \|a_K\|^2) =t $  and so the arguments given above imply that
\begin{align}
    D(T^{1/2}  \bu  +\bw  \, \| \, \bw) & 
    = D( \diag(t)^{1/2}  \bu  + \bw \, \| \, \bw ).
\end{align}

By the chain rule for relative entropy, this result extends to the setting where $U$ is an $n \times p$ random matrix with the property that each row has at most one entry that is not deterministically zero, i.e., we can write 
\begin{align}
    D(U T^{1/2}   +W  \, \| \, W) & 
    = D( U \diag(t)^{1/2}   + W \,  \| \, W ).
\end{align}
where $W$ is an independent $n \times p$ matrix with independent standard Gaussian entries.
Finally, the stated result for the matrix tensor product model follows from noting that both $X$ and $X \otimes X$ have the property that at most one entry in each row is not deterministically zero.
\end{proof}

In view of Lemma~\ref{lem:diagS}, the relation in \eqref{eq:cD_N_to_cDmpt_N} can simplified as 
\begin{align}
\cD_N(r,\Lambda) =  \cD^\mtp_N(\diag(r),\diag(\gvec(\Lambda)). 
\end{align}
Moreover, if $\tilde{R}$ is a $K \times K$ symmetric positive semidefinite matrix with diagonal given by $\tilde{r}$,  we have
\begin{align}
\cD^\mtp_N(\diag(r)  +  \tilde{R}, 0)&= \cD^\mtp_N(\diag(r_k  + \tilde{r})  , 0) = \sum_{k=1}^K \frac{n_k}{N} D_k(r_{k} + \tilde{r}_{k})
\end{align}
where the first step is due to Lemma~\ref{lem:diagS} and second step follows from Assumption~\ref{assumption:1}. Combining this expression with the fact that $\frac{1}{N} \ex{ X^\top X} = \diag(n_1/N, \dots, n_K/ N)$, leads to the following simplification of  approximation formula: 
\begin{align}
 \hat{\cD}^\mtp_N(\diag(r),\diag(\gvec(\Lambda)))
&=  \adjustlimits \max_{q  } \inf_{\tilde{r}} \left\{ \sum_{k=1}^K \frac{n_k}{N}  D_k(r +\tilde{r}_{k}) +\frac{1}{2} q^\top \Lambda q - \frac{1}{2} \tilde{r}^\top q  \right\}.  \label{eq:hatcD_N}
\end{align}
Here, the maximum is over $K$-dimensional vectors  satisfying $0 \le q_k \le n_k/N$ and the infimum is over $K$-dimensional vectors with non-negative entries. Finally, taking the $N \to \infty$ limit of this expression leads to the formula for the limit given in Theorem~\ref{thm:entropy}.

\subsection{Proof of Theorem~\ref{thm:mmse}} \label{app:thmmmse}

The proof of Theorem~\ref{thm:mmse} follows from fact that changes in the relative entropy with respect to the parameters $(r,\Lambda)$ can be related to the MMSE in estimating the underlying variables. As a consequence of the I-MMSE relationship \cite{guo:2005}, it can be shown that  $\cD_N$ is differentiable on the interior of its domain with  partial derivatives:
\begin{align}
\partial_{r_k} D_N(r, \Lambda) = \frac{1}{2N}  \ex{ \| \ex{ \bx_k \mid Y} \|_2^2}, \quad \partial_{\lambda_{k\ell}} D_N(r, \Lambda) = \frac{1}{2N^2}  \ex{ \| \ex{ \bx_k\bx_\ell^\top \mid Y} \|_F^2}.
\end{align}
These derivatives can be defined uniquely on the boundaries ($r_k=0$ or $\lambda_{kl}=0$) according to the right derivatives, which exist and are finite due the finite fourth moment assumption \cite[Proposition~7]{guo:2011}. The connection with the MMSE follows from the relationship
\begin{align}
    \ex{ \| \bx_k - \ex{ \bx_k \mid Y}\|^2} & = \ex{ \|\bx_k\|^2} - \ex{ \| \ex{ \bx_k \mid Y} \|_2^2}\\
     \ex{ \| \bx_k \bx_\ell  - \ex{ \bx_k\bx_\ell^\top \mid Y}\|_F^2} & = \ex{ \|\bx_k \bx_\ell^\top\|_F^2} - \ex{ \| \ex{ \bx_k \bx_\ell^\top \mid Y} \|_F^2}.
\end{align}
The assumption that the entries are independent with second moment equal to one means that $\ex{ \|\bx_k\|^2} = n_k$ and $\ex{ \|\bx_k \bx_\ell\|_F^2} = n_k n_\ell$ for $k \ne \ell$. For the diagonal blocks, we can write $\ex{ \|\bx_k \bx_k^\top\|_F^2} = n_k \xi_k  + n_k(n_k - 1)$
where $\xi_k$ is the fourth moment of $P_k$ and so $n_k^{-2} \ex{ \|\bx_k \bx_k^\top\|_F^2}$ converges to one in the $N \to \infty$ limit.

The next step is to relate the derivatives of  $\cD_N$ to the derivatives of its limit $\cD$. This is possible due to the fact that  $\cD_N$ is convex, which implies that $\cD$ is convex, and hence differentiable almost everywhere, and furthermore that the derivatives converge at every point where the limit is differentiable  \cite[Proposition 4.3.4]{hiriart-urruty:93}. 

With these results in hand, the asymptotic MMSE can be analyzed by studying the limit $\cD$. If for a given pair $(r,\Lambda)$ the maximum in the definition of $\cD$ is attained at a unique point $q^*$ then $\cD$ has partial derivatives with respect to the nonzero entries in $(r,\Lambda)$. Specifically, for   $r_k >0$ and $\lambda_{k\ell} >0$ it can be verified that
\begin{align}
\partial_{r_k} \cD(r, \Lambda) = \frac{1}{2} q^*_k,
\qquad \partial_{\lambda_{k\ell}} \cD(r, \Lambda) = \frac{1}{2}  q^*_{k}q_\ell^*. \label{eq:partial_cD}
\end{align}
For $r=0$ and $\lambda_{k\ell}= 0$ these expressions still hold for the right derivatives and it can be shown (see \cite[Section~II-B]{reeves:2020MTP}) that they provide one-sided bounds of the form
\begin{align}
\limsup_{N \to \infty} \frac{1}{N}  \ex{ \|  \ex{ \bx_k \mid Y} \|_2^2}  &\le q^*_k, \qquad 
\limsup_{N \to \infty} \frac{1}{N^2}  \ex{ \|  \ex{ \bx_k\bx_\ell^\top \mid Y} \|_F^2}  \le q^*_k q^*_\ell.
\end{align}
Noting that these are equalities  whenever the upper bound is equal to zero and combining with the arguments above leads to the conditions in Theorem~\ref{thm:mmse}.

\subsection{Almost Everywhere Uniqueness of Maximizer} \label{sec:uniqueness}

Recall that $\cD$ is convex and thus differentiable almost everywhere on the interior of its domain~\cite[Theorem~25.5]{rockafellar:1970}. For each point $(r, \Lambda)$ where $\cD$ is differentiable, we can combine the  envelop theorem \cite{milgrom:2002} with  \eqref{eq:partial_cD} to conclude that the maximum in \eqref{eq:cD} has a unique maximizer $q^*$. Finally, this argument can also be applied with respect to a subset of entries in $(r, \Lambda)$,  providing that the corresponding partial derivatives uniquely characterize $q^*$. For example, for any fixed $r$  and collection of off-diagonal terms $(\lambda_{k\ell} : k \ne \ell)$, the  maximizer is unique for almost all $\lambda_{11}, \dots, \lambda_{KK} \in (0,\infty)$.

\subsection{Proof of Theorem~\ref{thm:wpca}} \label{app:wpca}
Recall that the relative entropy function given in Theorem~\ref{thm:entropy} for the optimally weighted PCA setting described in Section~\ref{sec:WPCA} is given by
\begin{align} \label{eq:pcaare}
    \cD(0, \Lambda) = \adjustlimits \max_{0 \leq q \leq \beta} \inf_{\tilde{r} \ge 0} \left\{ \sum_{k=0}^L \beta_k D_k(\tilde{r}_k) + \frac{q_0}{2} \sum_{\ell = 1}^L{\frac{q_\ell}{\sigma_\ell^2}}
    - \frac{1}{2} \tilde{r}^\top q \right\}
\end{align}
where $D_k \, : \, [0, \infty) \to [0, \infty)$ is the single letter relative entropy function for $k = 0, \dots, L$. The assumption that the entries of $\bu$ and $\bv_\ell$ have independent standard Gaussian entries means that the single-letter relative entropy functions can be expressed in closed-form as $D_k(\tilde{r}_k) =\frac{1}{2}  (\tilde{r}_k - \log (1 + \tilde{r}_k))$ for $k = 0,1, \dots, L$. Observe that the objective function in \eqref{eq:pcaare} is convex in $\tilde{r}$. By  differentiation, one finds that for  $0 \leq q < \beta$, the inifmum is attained at a unique point $r^*= r^*(q)$ satisfying: 
\begin{align}
    r^*_k = \frac{ q_k}{ \beta_k - q_{k}}, \qquad k  = 0, 1, \dots, L
\end{align}
Plugging this value into \eqref{eq:pcaare} and noting that the objective is continuous in $q$ leads to the following expression:
\begin{align}
 \sup_{0 \leq q < \beta} \left\{ \frac{q_0}{2} \left( 1 + \sum_{\ell=1}^L{\frac{q_\ell}{\sigma_\ell^2}} \right) + \frac{1}{2} \sum_{\ell = 1}^L{q_\ell} - \frac{1}{2} \sum_{k = 0}^L{\beta_k \log(\beta_k)} + \frac{1}{2} \sum_{k = 0}^L{\beta_k \log(\beta_k - q_k)} \right\}\label{eq:pcaare_b}
\end{align}

This expression can be simplified further by noting that for fixed $0 \le q_0 < \beta_0$ the objective function in \eqref{eq:pcaare_b} is strictly concave in $(q_1, \dots, q_L)$ and thus the maximum is attained at the  unique point $(q_1^*(q_0), \dots, q_L^*(q_0))$ characterized by
\begin{align}
q^*_{\ell} = \frac{\beta_\ell q_0}{ \sigma_\ell^2 + q_0} , \qquad \ell = 1, \dots, L. \label{eq:q_star_ell}
\end{align}
In the case $q_0= 0$ this follows since the objective function is decreasing in $q_{\ell}$ for $\ell \ge 1$ and thus the maximum is attained at the boundary point $q^*_1 = \dots q^*_L=0$. In the case $q_0 > 0$ this follows from differentiation. Evaluating \eqref{eq:pcaare_b} at the point given by \eqref{eq:q_star_ell} and then simplifying terms leads to the following expression involving a univariate optimization problem: 
\begin{align} \label{eq:pcaare_c}
    \sup_{0 \le q_0 < \beta_0} \frac{1}{2}F(q_0) , \qquad F(x) \coloneqq x -  \beta_0 \log\left(\frac{\beta_0}{\beta_0 - x} \right) +   \sum_{\ell=1}^L  \left(  \frac{\beta_\ell x}{\sigma_\ell^2}    -  \beta_\ell \log\left(1 + \frac{x}{ \sigma_\ell^2} \right) \right)
\end{align}

The objective function $F$ is continuous on $[0,\beta_0)$ and converges to $- \infty$ at $x \nearrow \beta_0$. Hence, the maximum is attained attained on $[0,\beta_0)$. Furthermore, the derivative can be expressed as
\begin{align}
    F'(x) &= - \frac{ x}{\beta_0 - x}  + \sum_{\ell=1}^L \frac{\beta_\ell x}{ \sigma_\ell^2 (\sigma_\ell^2 + x)}. 
\end{align}
From inspection, it is clear that $F'$ has a root at $x = 0$. To explore the possible roots on $(0,\beta_0)$ define
\begin{align}
R(x) & \coloneqq \left( 1 -  \frac{ \beta_0}{x} \right) F'(x)  =  1   - \sum_{\ell=1}^L \frac{\beta_\ell (\beta_0 -x)}{ \sigma_\ell^2 (\sigma_\ell^2 + x)}  \label{eq:R}
\end{align}
Note that $(1- \beta_0 / x)$ is strictly negative on $(0, \beta_0)$ and so $R$ and $F'$ have the same roots and opposite signs over this interval.  

At this point we recognize that $R$ is the same as function appearing in  \citet[Theorem~6]{hong:2018}, adapted to the notation used in this paper. The analysis in \cite{hong:2018} shows that $R$ is strictly increasing over $(0, \beta_0)$ and converges to 1 as $x \nearrow \beta_0$. Consequently, it follows from \eqref{eq:R} that there are only two cases to consider:
\begin{itemize}
    \item If $R(0) \ge 0$ then $F'$ is strictly negative on $(0,\beta_0)$, and so the unique maximizer is $q_0^* = 0$. 
    
    \item Conversely, if $R(0) < 0$, then there exists a single root $x^*$ on $(0,\beta_0)$. Moreover, $F'$ is strictly positive on $(0,x^*)$ and strictly negative on $(x^*, \beta_0)$ and thus $q_0^*= x^*$ is the unique maximizer of $F$.
\end{itemize}

To conclude, observe the condition on $R(0)$ can be expressed equivalently as
\begin{align} \label{eq:wpcapt}
 R(0) < 0    \quad \iff \quad   \sum_{\ell = 1}^L{\frac{\beta_0 \beta_\ell}{\sigma_\ell^4}} > 1
\end{align}
The above establishes Theorem~\ref{thm:wpca} and now we can use it in conjunction with Theorem~\ref{thm:mmse} and Theorem~\ref{thm:bounds} to comment on the optimality of WPCA in this setting. First recall that \citet[Theorem~6]{hong:2018} states that
\begin{align}
    \frac{\langle \hat{\bu}, \bu \rangle^2}{\|\bu\|^2 \|\hat{\bu}\|^2} \stackrel{a.s.} \longrightarrow q_0^{\text{WPCA}}/\beta_0
\end{align}
where $\bu, \hat{\bu} \in \reals^d$ and $\hat{\bu}$ is the estimate produced by WPCA with optimal weights. It should be noted that \cite{hong:2018} assume that $\bu$ is a unit vector, however we take it to be a vector with IID entries with zero mean and unit variance. This discrepancy leads to some additional normalization terms in the formulas presented here.

We consider the MSE in estimating $\bu \bu^\top$ and allow for the optimal scaling of $\hat{\bu}$. We recall that the formula for the MSE in this case was derived in Section~\ref{sec:methods} and is
\begin{align}
\inf_{\rho_u \ge 0 } \frac{1}{d^2}  & \ex{ \left\| \bu \bu^\top -  \rho_u^2 \frac{\hat{\bu} \hat{\bu}^\top}{\|\hat{\bu}\|^2} \right \|_F^2} = \frac{1}{d^2}\ex{  \|\bu\|^4}  - \frac{1}{d^2} \ex{ \frac{ \langle \bu, \hat{\bu}\rangle^2}{\|\hat{\bu}\|^2}}^2.
\end{align}
We can consider the above MSE in the asymptotic limit (i.e., as $N \to \infty$) and note that in this limit $\|\bu\|^2$ concentrates to a deterministic value of $d$. So, the asymptotic MSE for WPCA is given by
\begin{align}
    \lim_{N \to \infty}\frac{1}{d^2} \left( \ex{\|\bu\|^4} - \left(\ex{\frac{\langle \bu, \hat{\bu} \rangle^2}{\|\hat{\bu}\|^2}}\right)^2 \right) = 1 - \lim_{N \to \infty}\left( \ex{\frac{\langle \bu, \hat{\bu} \rangle^2}{\|\bu\|^2 \|\hat{\bu}\|^2}} \right)^2 = 1 - \left( \frac{q_0^{\mathrm{WPCA}}}{\beta_0} \right)^2
\end{align}
where the final equality holds because
\begin{align}
    \frac{\langle \bu, \hat{\bu} \rangle^2}{\|\bu\|^2 \|\hat{\bu}\|^2}
\end{align}
is a positive and bounded quantity. Therefore, almost sure convergence implies convergence in $L_1$. We also note that this MSE is lower bounded by the MMSE. Furthermore, from Theorem~\ref{thm:bounds}, the MMSE of $\bu \bu^\top$ is lower bounded by
\begin{align}
    1 - \left( \frac{q_0^*}{\beta_0} \right)^2
\end{align}
where $q_0^*$ is the result given in Theorem~\ref{thm:wpca} and is equal $q_0^{\mathrm{WPCA}}$ at and after the weak recovery condition is satisfied. Given that we have demonstrated an algorithm that attains a lower bound on the MMSE, we can conclude that the MSE of the optimally scaled WPCA estimator is equal to the MMSE. Hence, it can be said that WPCA achieves the fundamental limits in this setting.

\section{AMP for Heteroskedastic Rank-One Estimation} \label{app:AMP}

This section considers approximate inference methods for the heteroskedastic rank-one observation model given by
\begin{align}
    y_{ij} = \sqrt{ \frac{\lambda_{ij}}{N}} x_i x_j  + w_{ij}, \qquad  i,j = 1, \dots, n \label{eq:yij} 
\end{align}
where $(x_i)$ are unknown random variables,  $(w_{ij})$ are independent standard Gaussian noise terms, and $(\lambda_{ij})$  and $N$ are known parameters. Furthermore, it is assumed that the variables $(x_i)$ are  independently (but not identically) distributed with $x_i \sim P_i$ for $i =1, \dots, n$. Note that the groupwise model \eqref{eq:Ykl} corresponds to the special case where the parameters $(\lambda_{ij})$  and the distributions are constant within groups. 

\paragraph{Symmetrization and Prepossessing}
To simplify the exposition of the algorithms it is  convenient to apply a prepossessing step that re-scales and symmetrizes the observations. Specifically, we define the modified observations $(\tilde{y}_{ij})$ and parameters $(\tilde{\lambda}_{ij})$ according to
\begin{align}
\tilde{y}_{ij} & = \begin{dcases}  \sqrt{\lambda_{ij}} \,  y_{ij} + \sqrt{\lambda_{ji}} \,  y_{ji} & i \ne j \\
\sqrt{ \lambda_{ii}}\,  y_{ii}, & i=j
\end{dcases}, \qquad 
\tilde{\lambda}_{ij}   =
\begin{dcases}
\lambda_{ij} + \lambda_{ji} , & i \ne j\\
\lambda_{ii} & i = j
\end{dcases}
\end{align}
By the orthogonal invariance of the standard Gaussian distribution it follows that $(\tilde{y}_{ij})$ are sufficient statistics for estimation of $(x_i)$. Furthermore, the distribution of $(\tilde{y}_{ij})$ is given by the symmetric model
\begin{align}
    \tilde{y}_{ij} = \frac{1}{ \sqrt{N}} \tilde{\lambda}_{ij} x_i x_j  + \tilde{\lambda}^{1/2}_{ij} \tilde{w}_{ij} \label{eq:yijsym}
\end{align}
where $\tilde{w}_{ij} = \tilde{w}_{ji}$ with $( \tilde{w}_{ij} \, :\, 1 \le i \le j \le n)$ independent standard Gaussian variables. Using this parameterization, the likelihood function is proportional to 
\begin{align}
\prod_{1 \le i \le j \le n}\exp\left\{  \frac{ \tilde{y}_{ij}x_i x_j}{\sqrt{N}}  -   \frac{ \tilde{\lambda}_{ij}  x_i^2 x_j^2 }{2N}  \right\}. \label{eq:yijsym_likelihood}
\end{align}

\paragraph{Belief Propagation}  We begin with a derivation of the belief propagation algorithm \cite{pearl:1998}. To simplify the derivation we assume that each probability measure $P_i$ had a density $f_i$ with respect to the Lebesgue measure. In view of \eqref{eq:yijsym_likelihood}, the conditional distribution of the variables $(x_i)$ given the  observations  $(\tilde{y}_{ij} )$ has a density that is proportional to 
\begin{align}
    \prod_{i = 1}^n{\phi_i(x_i)} \prod_{1 \leq i < j \leq n}\psi_{ij}\left( x_i, x_j \right),  \label{eq:px_factor}
\end{align}
where 
\begin{align}
\phi_i(x_i) & \coloneqq f_i(x_i) \exp\left\{  \frac{ \tilde{y}_{ii} x_i^2}{\sqrt{N}} - \frac{\tilde{\lambda}_{ii}  x_i^4}{2 N} \right \}, \qquad 
\psi_{ij}\left( x_i, x_j\right)  \coloneqq \exp\left\{  \frac{ \tilde{y}_{ij}x_i x_j}{\sqrt{N}}  -   \frac{ \tilde{\lambda}_{ij}  x_i^2 x_j^2 }{2N}  \right\}
\end{align}

The belief propagation algorithm is described by a collection of messages $(\mu_{ij}\, : \,  1 \le  i , j \le n, \; i \ne j)$ where each  $\mu_{ij}$  is a non-negative function that describes the `influence' of variable $i$ on variable $j$. Starting at a given initialization, these messages are determined recursively via the update rule
\begin{align}
    \mu_{ij}(x_j) \leftarrow  \int \phi_i(x_i) \psi_{ij}(x_i, x_j)  \prod_{k \in [n] \backslash \{i , j\} }  \mu_{ki}(x_i) \, d x_i, \label{eq:mu_ij}
\end{align}
where $[n] = \{1\dots, n\}$. The approximation to the marginal of the $i$-th variable associated with given collection of messages is given by the probability density function:
\begin{align}
x_i \mapsto   \frac{\phi_i(x_i) \prod_{ j \in [n] \backslash i} \mu_{j i}(x_i)}{ \int \phi_i(x'_i) \prod_{ j \in [n] \backslash i} \mu_{j i}(x'_i)\, d x_i'}. \label{eq:bp_mparginal}
\end{align}
Note that this final approximation is invariant to rescaling of the messages, and thus at each stage in the algorithm the message $\mu_{ij}$ can be rescaled by an arbitrary positive constant.

\paragraph{Relaxed Belief Propagation} Next we consider a relaxed belief propagation algorithm that is obtained by an approximation of the update rule. The first step in this approximation is to replace the function $\psi_{ij}(x_i,x_j)$ by its second order expansion with respect to the term  $(x_i x_j)/\sqrt{N}$ evaluated at zero:
\begin{align}
    \psi_{ij}(x_i,x_j) \approx 1  + \frac{\tilde{y}_{ij} x_i x_j}{\sqrt{N}} + \frac{1}{2}  \frac{ (\tilde{y}_{ij}^2 -\tilde{\lambda}_{ij})  x^2_i x^2_j}{N}.
\end{align}
Plugging this approximation into \eqref{eq:mu_ij} leads to the following integral
\begin{align}
  \int \phi_i(x_i)  \left( 1  + \frac{\tilde{y}_{ij} x_i x_j}{\sqrt{N}} + \frac{1}{2}  \frac{ (\tilde{y}_{ij}^2 - \tilde{\lambda}_{ij} )  x^2_i x^2_j}{N} \right)  \prod_{k \in [n] \backslash \{i , j\} }  \mu_{ki}(x_i) \, d x_i \label{eq:message_taylor_approx}
\end{align}
To characterize the solution to this integral, define
\begin{align}
m_{ij} & = \frac{\int x_i \phi_i(x)   \prod_{k \in [n] \backslash \{i , j\} }  \mu_{ki}(x_i) \, d x_i}{\int \phi_i(x')   \prod_{k \in [n] \backslash \{i , j\} }  \mu_{ki}(x'_i) \, d x'_i} \\
v_{ij} &= \frac{\int x^2_i \phi_i(x)   \prod_{k \in [n] \backslash \{i , j\} }  \mu_{ki}(x_i) \, d x_i}{\int \phi_i(x')   \prod_{k \in [n] \backslash \{i , j\} }  \mu_{k\to i}(x'_i) \, d x'_i}  - \hat{x}^2_{ij}
\end{align}
to be the mean and variance, respectively, of the probability density function 
\begin{align}
   x_i \mapsto  \frac{ \phi_i(x)   \prod_{k \in [n] \backslash \{i , j\} }  \mu_{ki}(x_i) }{\int \phi_i(x')   \prod_{k \in [n] \backslash \{i , j\} }  \mu_{ki}(x'_i) \, d x'_i}. \label{eq:marginal_ij}
\end{align}
This density is similar to the marginal approximation in \eqref{eq:bp_mparginal} except that message from the $j$-th variable is also excluded. Using this notation, the integral in \eqref{eq:message_taylor_approx} is proportional to 
\begin{align}
    1  + \left( \frac{\tilde{y}_{ij} m_{ij} }{\sqrt{N}} \right) x_j   + \frac{1}{2}\left( \frac{(\tilde{y}^2_{ij} - \tilde{\lambda}_{ij})  \left(m_{ij}^2 + v_{ij} \right) }{N} \right) x_j^2.
\end{align}
Using the approximation
$1 + a u + \frac{1}{2}(a^2 -b) u^2 \approx \exp( a u - \frac{1}{2} b u^2)$ for $u \approx 0$ and recalling that we can rescale the message $\mu_{ij}$ by an arbitrary positive constant leads to the modified update rule
\begin{align}
\mu_{ij}(x_j) \leftarrow   \exp\left\{ a_{ij}  x_j   - \frac{1}{2} b_{ij} x_j^2\right\}  \label{eq:muij_approx}
\end{align}
where 
\begin{align}
    a_{ij} &= \frac{\tilde{y}_{ij} m_{ij} }{\sqrt{N}}, \qquad 
    b_{ij} =\frac{ \tilde{\lambda}_{ij} m_{ij}^2 + (\tilde{\lambda}_{ij} - \tilde{y}_{ij}^2)  v_{ij}  }{N}. \label{eq:expcoeffs}
\end{align}
According to the update rule \eqref{eq:muij_approx}, the probability density in \eqref{eq:marginal_ij} is defined by the (unnormalized) density
\begin{align}
 \phi_i(x_i) \prod_{ k \in [n] \backslash \{i, j\}} \exp\left\{ a_{ki}  x_i   - \frac{1}{2} b_{ki} x_i^2\right\}  = \phi_i(x_i)  \exp\left\{ \sum_{ k \in [n] \backslash \{i,j\} }  a_{ki}  x_i   - \frac{1}{2} b_{ki} x_i^2\right\} 
\end{align}
Here, we see that the products of the messages are described compactly in terms of a linear combination of the $a_{ki}$ and $b_{ki}$ terms.

The relaxed belief propagation can be described as follows. For $i =1, \dots,n$ define the function
\begin{align}
    \eta_i(a, b) = \frac{  \int x_i  \exp \left\{ a x_i - \frac{1}{2}bx_i^2 \right\} \phi_i(x_i)\, d x_i }{ \int \exp \left\{ ax'_i - \frac{1}{2} b (x'_i)^2 \right\} \phi_i(x'_i) dx'_i}  \label{eq:etai}
\end{align}
and let $\eta'_i(a,b)$ denote the partial derivative with respect to the first argument. Note that $\eta_i(a,b)$ and $\eta'_i(a,b)$ represent the mean and variance, respectively, of the (unormalized) density  $x_i \mapsto \phi_i(x_i) \exp( a x_i - \frac{1}{2} b x_i^2)$.  Starting with initial values for $(m_{ij},v_{ij})$ the updates are defined according to
\begin{subequations}
\label{eq:bprelaxed}
\begin{align}
  m_{ij} 
   & \leftarrow \eta_i\left( \sum_{k \in [n]\backslash \{i,j\}} a_{ki} ,\sum_{k \in [n]\backslash \{i,j\}} b_{ki}   \right) \\
  v_{ij}
   & \leftarrow \eta'_i\left( \sum_{k \in [n]\backslash \{i,j\}} a_{ki} ,\sum_{k \in [n]\backslash \{i,j\}} b_{ki}   \right)
\end{align}
\end{subequations}
where the terms $(a_{ki}, b_{ki})$ are defined with respect to the current values of $(m_{ki}, v_{ki})$ according to \eqref{eq:expcoeffs}. 

\paragraph{Approximate Message Passing} One of the limitations of the relaxed belief propagation algorithm in \eqref{eq:bprelaxed} is that it requires keeping track of $n^2$ terms.  AMP  can be viewed as an approximation to the relaxed belief propagation algorithm that requires only $n$ terms.

Before stating the AMP algorithm, it is useful to consider the version of relaxed belief propagation where all of the updates are made in parallel. Specially, starting with an initialization $(m_{ij}^0, v_{ij}^0)$, the updates at time $t = 1 ,2,\dots$ are given by
\begin{subequations}
\begin{align}
a^t_{ij} &= \frac{\tilde{y}_{ij}}{\sqrt{N}} m_{ij}^t\\
b^t_{ij} & = \frac{\tilde{\lambda}_{ij}}{N} (m_{ij}^t)^2  + \frac{ \tilde{\lambda}_{ki} - \tilde{y}_{ij}}{N} v^t_{ij}\\
  m^{t+1}_{ij} 
   & = \eta_i\left( \sum_{k \in [n]\backslash \{i,j\}} a^t_{ki} ,\sum_{k \in [n]\backslash \{i,j\}} b^t_{ki}   \right) \\
  v^{t+1}_{ij}
   & = \eta'_i\left( \sum_{k \in [n]\backslash \{i,j\}} a^t_{ki} ,\sum_{k \in [n]\backslash \{i,j\}} b^t_{ki}   \right)
\end{align}
\end{subequations}
In comparison, the corresponding version of AMP is given by
\begin{subequations}
\label{eq:ampij}
\begin{align}
a^t_{i} & = \frac{1}{\sqrt{N}} \sum_{k \in [n]\backslash i} \tilde{y}_{ki} m_{k}^t  -  \frac{1}{N}  m_{i}^{t-1}\sum_{k \in [n]\backslash i} \tilde{y}_{ki}  v_k^t  \tilde{y}_{ik} \\
b^t_{i} & =  \frac{1}{N} \sum_{k \in [n]\backslash i}\left(\tilde{\lambda}_{ki} (m_{k}^t)^2  + ( \tilde{\lambda}_{ki} - \tilde{y}^2_{ki}) v^t_{k} \right)\\
    m_i^{t+1}  &= \eta_i \left( a_i^t, b_i^t \right) \\
v_{i}^{t+1} &= \eta'_i \left( a_i^t, b_i^t \right)
\end{align}
\end{subequations}
The relationship between these algorithms can be derived heuristically based on a decomposition of the means of the form $m_{ij}^t = m_i^t + \eps_{ij}^t $ where $m_i^t$ does not depend on $j$ and $\eps_{ij}^t$ is a small fluctuation; see e.g.,~\cite[Appendix~A]{bayati:2011}.

To express the AMP algorithm using vector notation, we define $\bbm^t$ and $\bv^t$ be $n \times 1$ vectors containing the means and variances, respectively, at iteration $t$. Also, we define $\tilde{Y}$ and $\tilde{\Lambda}$ to be $n \times n$ symmetric matrices with diagonal entries equal to zero and off-diagonal entries given by $\tilde{y}_{ij}$ and $\tilde{\lambda}_{ij}$, respectively. The AMP algorithm in \eqref{eq:ampij} can be written compactly as
\begin{subequations}
\label{eq:amp_full}
\begin{align}
\ba^t & = \frac{1}{\sqrt{N}} \tilde{Y} \bbm^t  -  \frac{1}{N}     (\tilde{Y}^{\circ2}  \bv^t ) \circ \bbm^{t-1}\\
\bb^t & =  \frac{1}{N} \tilde{\Lambda} (\bbm^t)^{\circ 2}  + \frac{1}{N} ( \tilde{\Lambda} - \tilde{Y}^{\circ 2}) \bv^t \\
    \bbm^{t+1}  &= \eta \left( \ba^t, \bb^t \right) \\
    \bv^{t+1} &= \eta' \left( \ba^t, \bb^t \right)
\end{align}
\end{subequations}
where $\eta$ and $\eta'$ are obtained by stacking the functions $\eta_i$ and $\eta_i'$, respectively, for $i = 1, \dots, n$.

\begin{remark}
In the AMP algorithm given in \eqref{eq:amp_full} The dependence on distribution of $x_i$ is encapsulated by the function $\eta_i(a,b)$ defined in \eqref{eq:etai}. This means that the algorithm can be applied in the case where $x_i$ is drawn according to a probability measure $P_i$ provided that the functions
\begin{align}
(a,b) \mapsto    \int x^p_i  \exp \left\{ a\,  x_i - \frac{1}{2}b\, x_i^2 \right\} \exp\left\{  \frac{ \tilde{y}_{ii} x_i^2}{\sqrt{N}} - \frac{\tilde{\lambda}_{ii}  x_i^4}{2 N} \right \}  \, d P_i(x_i), \quad p = 0, 1,2 
\end{align}
can be approximated numerically. 
\end{remark}

\begin{remark}\label{rem:etai}
If the parameter $\lambda_{ii}$ is small relative to the global scale parameter $N$, then the data point $\tilde{y}_{ii}$ has negligible impact and the term $\phi_i(x_i)$ in \eqref{eq:etai} can be replaced by the density of $x_i$. This is precisely  what happens in the usual rank-one estimation model where all the $\lambda_{ij}$ are identical. By contrast, the heteroskatistic model \eqref{eq:yij} allows for the possibility that the diagonal terms $\lambda_{ii}$ are of the same order as $N$ and in this case, the diagonal terms should not be discarded.
\end{remark}

\paragraph{AMP for Groupwise Model} The AMP presented in Section~\ref{sec:AMP} is obtained by specializing  the AMP algorithm in \eqref{eq:amp_full} to the to setting of the groupwise model. Following the discussion in Remark~\ref{rem:etai} the function $\eta_i(a,b)$ defined in \eqref{eq:etai} is replaced by the definition given in \eqref{eq:etak}, which omits the dependence on the diagonal entries of the observations.

\section{Numerical Approximation of Relative Entropy and MMSE} \label{app:MMSEApprox}
Numerically approximating the relative entropy and MMSE requires solving a low-dimensional saddle point problem and generating expressions for the single-letter relative entropy and its first derivative. In this work, we only consider numerical approximation of the formulas for the two-group case, as in \eqref{eq:twogroup}. However, the single-letter relative entropy and its first derivative will be the same in the $K$-group case and our strategy for estimating the global optimum of the saddle point problem would be a reasonable strategy for $K > 2$ as well.

\subsection{Estimation of the Global Optimum}
In the two-group setup, we only have to perform an optimization over four scalar variables in order to obtain the relative entropy and MMSE. To further simplify the process, we can note that the envelope theorem (see \cite{milgrom:2002}) gives us that we need only consider the stationary points of the objective. At first glance of \eqref{eq:twogprob}, this is far from obvious. In practice, we use a multivariate root finding algorithm from the SciPy \cite{scipy:2001} optimization package in order to find the locations where the gradient is equal to zero.

\subsection{Approximation of the Single-letter Relative Entropy}
In order to deal with saddle point problem in \eqref{eq:twogprob}, we need to obtain a formula for the single-letter relative entropy, defined in \eqref{eq:slre}, and its first derivative. In this work, we consider standard Gaussian priors and discrete priors. In the case of standard Gaussian priors, we have $D_{k}(\gamma) = \frac{1}{2}( \gamma - \log(1 + \gamma))$ and $\partial_{\gamma} D_k(\gamma) = \frac{1}{2} \frac{\gamma}{1 + \gamma}$. We also consider discrete priors which have the following form
\begin{equation} \label{DiscretePdf}
    P_X = \sum_{i=1}^M p_i \delta_{a_i}
\end{equation}
where $\delta_{a_i}$ denotes a point mass function at $a_i$ and $p_i$ are probabilities for $i = 1,..,M$ and $P_X$ denotes the probability measure. In order to obtain the single-letter relative entropy, we consider the random variable $Y = \sqrt{\gamma} X + Z$, where $X \sim P_X$ and $Z \sim \mathcal{N}(0, 1)$ are independent. The probability density function of $Y$ is a Gaussian mixture, which is given by
\begin{equation}
    f_Y(y) = \sum_{i = 1}^M{p_i \mathcal{N}(y; \sqrt{\gamma}a_i, 1)}
\end{equation}
where $\mathcal{N}(y; \mu, \sigma^2)$ denotes the Gaussian probability density function with mean $\mu$ and variance $\sigma^2$ evaluated at $y$. The single-letter relative entropy function and its derivative can be expressed as one-dimensional Gaussian integrals, which are approximated numerically.

\section{Additional Figures} \label{app:addfigs}
\begin{figure}
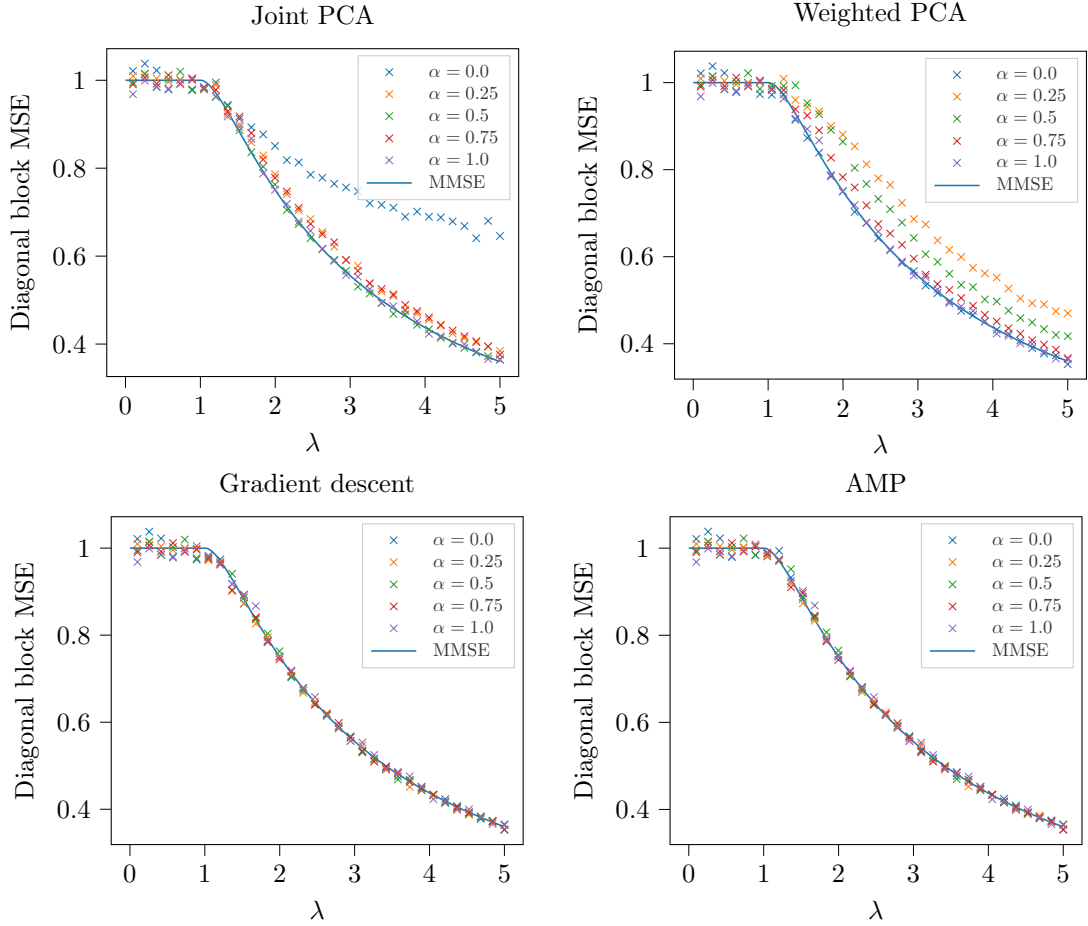

\centering
    \begin{subfigure}{.45\textwidth}
      \centering
      \input{figures/fig2a.tex}
    \end{subfigure}
    \begin{subfigure}{.45\textwidth}
      \centering
      \input{figures/fig2b.tex}
    \end{subfigure}
    \begin{subfigure}{.45\textwidth}
        \centering
        \input{figures/fig2c.tex}
    \end{subfigure}%
    \begin{subfigure}{.45\textwidth}
        \centering
        \input{figures/fig2d.tex}
    \end{subfigure}
    \caption{\label{fig:GaussResults2} MSE in estimating the rank-one matrices $\bu\bu^\top$ and $\bv \bv^\top$ in the two-group model \eqref{eq:twogroup} as function of $\lambda$ for various $\alpha$ with $n = 512$ and IID standard Gaussian priors. The asymptotic MMSE (solid line) is given by Theorem~\ref{thm:mmse}. The subplots show the diagonal block MSE of joint PCA (upper left), weighted PCA (upper right), gradient descent (lower left), and AMP (lower right).}
\end{figure}

\paragraph{Gaussian Priors} The case where both $\bu$ and $\bv$ have independent standard Gaussian entries is shown in Figure~\ref{fig:GaussResults2}. In this setting, the asymptotic MMSE is constant for all $0 \le \alpha \le 1$ and the weak detection threshold (i.e., the smallest value of $\lambda$ such that the asymptotic MMSE is strictly less than the  MMSE obtained without any observations) occurs at the critical threshold $\lambda^* =1$. The empirical performance of both gradient descent and AMP shows excellent agreement with the formulas for the asymptotic MMSE. By contrast, the spectral methods are sensitive to the choice of the  parameter $\alpha$. More specifically, the spectral methods  perform well at the special values of $\alpha$ for which the two-group model reduces to one of the previously studied spiked matrix models. (These values are given by $\alpha = 1/2$ for joint PCA and $\alpha \in\{0,1\}$ for weighted PCA.) However, as $\alpha$ deviates from these special values the performance degrades sharply, indicating that these spectral methods are sub-optimal in general.

\paragraph{Non-Gaussian Priors} Two examples involving non-Gaussian priors are shown in Figure~\ref{fig:NGAMP}. In the first example, both $\bu$ and $\bv$ have independent Rademacher entries and in the second example, $\bu$ has independent Bernoulli(0.1) entries (shifted and scaled to mean zero  and variance one) and $\bv$ has independent standard Gaussian entries. In both cases, empirical results are shown only for AMP. The asymptotic MMSE is invariant to $\alpha$ in the first example, where $\bu$ and $\bv$ have the same distribution, but depends on $\alpha$ in the second example, where $\bu$ and $\bv$ have different distributions. Interestingly, the MMSE curves in the second example are ordered for both the MMSE of $\bu \bu^\top$ and $\bv \bv^\top$, however, the order is reversed between the two. In particular, smaller values of $\alpha$ result in better performance (lower MMSE) in estimating $\bu \bu^\top$, but have worse performance (higher MMSE) in estimating $\bv \bv^\top$.

\begin{figure}[t]
\centering
    \begin{subfigure}{.45\textwidth}
        \input{figures/fig3a.tex}
    \end{subfigure} \\
    \begin{subfigure}{.45\textwidth}
        \input{figures/fig3b.tex}
    \end{subfigure}
    \begin{subfigure}{.45\textwidth}
        \input{figures/fig3c.tex}
    \end{subfigure}
    \caption{MSE in estimating the rank-one matrices $\bu\bu^\top$ and $\bv \bv^\top$ in the two-group model \eqref{eq:twogroup} as function of $\lambda$ for various $\alpha$ with $n = 512$. The asymptotic MMSE (solid line) is given by Theorem~\ref{thm:mmse}. The empirical MSE corresponds to AMP and is averaged over 32 Monte Carlo trials. In the top subplot, both $\bu$ and $\bv$ are Radamacher. In the bottom subplots, $\bu$ has a Bernoulli(0.1) prior (scaled and shifted to zero mean and unit variance) and $\bv$ is standard Gaussian. In the lower left subplot, we are plotting the MMSE and MSE in estimating $\bu \bu^\top$ and in the lower right subplot, we are plotting the MMSE and MSE in estimating $\bv \bv^\top$.} \label{fig:NGAMP}
\end{figure}
\end{document}

%% file: figures/fig1a.tex
\begin{tikzpicture}

\definecolor{color0}{rgb}{0.75,0,0.75}

\begin{axis}[
legend cell align={left},
legend style={
  fill opacity=0.7,
  draw opacity=1,
  text opacity=0.7,
  draw=white!80!black,
  nodes={scale=0.8}
},
tick align=outside,
tick pos=left,
x grid style={white},
xlabel={\(\displaystyle \alpha\)},
xmin=-0.05, xmax=1.05,
xtick style={color=white!33.3333333333333!black},
y grid style={white},
ylabel={Diagonal block MSE},
ymin = 0.6,
ymax = 1, 
ytick style={color=white!33.3333333333333!black},
width=0.9\linewidth,
height=0.7\textwidth
]
\addplot [draw=blue, fill=blue, mark options={rotate=180}, mark=triangle*, only marks]
table{%
x  y
0 0.852803718085915
0.032258064516129 0.927829224741182
0.0645161290322581 0.887087565823744
0.0967741935483871 0.863278394222514
0.129032258064516 0.839151934933108
0.161290322580645 0.823933323961253
0.193548387096774 0.809765972936673
0.225806451612903 0.801742121261805
0.258064516129032 0.782765768421329
0.290322580645161 0.768285781768067
0.32258064516129 0.763551783555805
0.354838709677419 0.758120002403725
0.387096774193548 0.764573664181008
0.419354838709677 0.750465561901657
0.451612903225806 0.756456708710604
0.483870967741935 0.746588587670564
0.516129032258065 0.747330007615125
0.548387096774194 0.752548667274953
0.580645161290323 0.75220445411237
0.612903225806452 0.756989040543046
0.645161290322581 0.762909955061221
0.67741935483871 0.768904593279413
0.709677419354839 0.77580886549242
0.741935483870968 0.772224011112316
0.774193548387097 0.797800322604628
0.806451612903226 0.807489934424504
0.838709677419355 0.823554068389202
0.870967741935484 0.842429743154037
0.903225806451613 0.856296334261048
0.935483870967742 0.89436852153084
0.967741935483871 0.928780186358947
1 0.749452285190167
};
\addlegendentry{Joint PCA}
\addplot [draw=green!50!black, fill=green!50!black, mark=triangle*, only marks]
table{%
x  y
0 0.748413388340015
0.032258064516129 0.767358622185749
0.0645161290322581 0.781106264337864
0.0967741935483871 0.807119873447101
0.129032258064516 0.815341914384437
0.161290322580645 0.83611178313809
0.193548387096774 0.856219067187561
0.225806451612903 0.890457140346005
0.258064516129032 0.889853988925681
0.290322580645161 0.908155300690037
0.32258064516129 0.91660460177724
0.354838709677419 0.93400988253944
0.387096774193548 0.946569604778583
0.419354838709677 0.919996404202431
0.451612903225806 0.924791743293649
0.483870967741935 0.898982995279284
0.516129032258065 0.880671599292015
0.548387096774194 0.869623041357738
0.580645161290323 0.856357959902114
0.612903225806452 0.840670876912366
0.645161290322581 0.82826257080363
0.67741935483871 0.82483661092879
0.709677419354839 0.809336161561581
0.741935483870968 0.792391211090317
0.774193548387097 0.79407981114117
0.806451612903226 0.788815877481936
0.838709677419355 0.77847341377299
0.870967741935484 0.772879625789331
0.903225806451613 0.765756754067728
0.935483870967742 0.761887013062856
0.967741935483871 0.753597181107027
1 0.749452285189643
};
\addlegendentry{Weighted PCA}
\addplot [draw=red, fill=red, mark=square*, only marks, mark size=1.5pt]
table{%
x  y
0 0.749083738675829
0.032258064516129 0.753190998568766
0.0645161290322581 0.750647960975437
0.0967741935483871 0.755971467193837
0.129032258064516 0.749282776399063
0.161290322580645 0.749801141673228
0.193548387096774 0.751196358236353
0.225806451612903 0.760238168496827
0.258064516129032 0.748714034804201
0.290322580645161 0.744525061977454
0.32258064516129 0.745667061545159
0.354838709677419 0.747501095182572
0.387096774193548 0.758124097987293
0.419354838709677 0.747233072551218
0.451612903225806 0.754525391214128
0.483870967741935 0.746348407767993
0.516129032258065 0.747258937411494
0.548387096774194 0.751815234714533
0.580645161290323 0.749284844213008
0.612903225806452 0.750231040647313
0.645161290322581 0.751976147337153
0.67741935483871 0.752945067651255
0.709677419354839 0.750048195230805
0.741935483870968 0.742717979162061
0.774193548387097 0.753110547697054
0.806451612903226 0.753784697065492
0.838709677419355 0.751321207458216
0.870967741935484 0.748896707230957
0.903225806451613 0.748253353278178
0.935483870967742 0.750857836311221
0.967741935483871 0.74954862957472
1 0.749452284147696
};
\addlegendentry{Gradient Descent}
\addplot [draw=color0, fill=color0, mark=x, only marks, mark size=3.5pt]
table{%
x  y
0 0.749526867459146
0.032258064516129 0.764767999258215
0.0645161290322581 0.754821881936236
0.0967741935483871 0.756876371223388
0.129032258064516 0.750242062042309
0.161290322580645 0.750362429916987
0.193548387096774 0.751553865170731
0.225806451612903 0.761165638433406
0.258064516129032 0.749536050217416
0.290322580645161 0.744932081507503
0.32258064516129 0.746493185251275
0.354838709677419 0.74843611425574
0.387096774193548 0.758609063475791
0.419354838709677 0.747260663462033
0.451612903225806 0.755412065284228
0.483870967741935 0.746864230790071
0.516129032258065 0.74792049415687
0.548387096774194 0.752118747656604
0.580645161290323 0.749821010286952
0.612903225806452 0.750361073917689
0.645161290322581 0.75229535244164
0.67741935483871 0.753389999624446
0.709677419354839 0.750353397882602
0.741935483870968 0.743033929783688
0.774193548387097 0.75322979071467
0.806451612903226 0.75382335154409
0.838709677419355 0.750869837166178
0.870967741935484 0.748667988037386
0.903225806451613 0.747820103238409
0.935483870967742 0.753506039886445
0.967741935483871 0.764173987618985
1 0.74945228511559
};
\addlegendentry{AMP}
\path [draw=black, semithick]
(axis cs:0,0.75)
--(axis cs:1,0.75);

\end{axis}

\end{tikzpicture}

%% file: figures/fig1b.tex
\begin{tikzpicture}

\definecolor{color0}{rgb}{0.75,0,0.75}

\begin{axis}[
legend cell align={left},
legend style={
  nodes = {scale=0.8},
  fill opacity=0.7,
  draw opacity=1,
  text opacity=0.7,
  draw=white!80!black
},
tick align=outside,
tick pos=left,
x grid style={white},
xlabel={\(\displaystyle \alpha\)},
xmin=-0.05, xmax=1.05,
xtick style={color=white!33.3333333333333!black},
y grid style={white},
ylabel={Diagonal block MSE},
ymin = 0.6,
ymax = 1, 
ytick style={color=white!33.3333333333333!black},
width=0.9\linewidth,
height=0.7\textwidth
]
\addplot [draw=blue, fill=blue, mark options={rotate=180}, mark=triangle*, only marks]
table{%
x  y
0 0.869321550003762
0.032258064516129 0.929619425082878
0.0645161290322581 0.892745716529593
0.0967741935483871 0.864685951700167
0.129032258064516 0.84221745115032
0.161290322580645 0.819860399915339
0.193548387096774 0.806924213876388
0.225806451612903 0.793463339458485
0.258064516129032 0.782753221527105
0.290322580645161 0.770477643533665
0.32258064516129 0.769979317769986
0.354838709677419 0.75477051206325
0.387096774193548 0.761312334417336
0.419354838709677 0.756637633620824
0.451612903225806 0.752223855269049
0.483870967741935 0.75319277554532
0.516129032258065 0.754136336982535
0.548387096774194 0.751270667096233
0.580645161290323 0.757894144158097
0.612903225806452 0.758410877111112
0.645161290322581 0.762363298758774
0.67741935483871 0.765632127847095
0.709677419354839 0.778999850649051
0.741935483870968 0.779443219904936
0.774193548387097 0.79278055787052
0.806451612903226 0.802667275650082
0.838709677419355 0.811721319733481
0.870967741935484 0.841011668751752
0.903225806451613 0.85864661085675
0.935483870967742 0.890879451024705
0.967741935483871 0.934454237052838
1 0.747840993339329
};
\addlegendentry{Joint PCA}
\addplot [draw=green!50!black, fill=green!50!black, mark=triangle*, only marks]
table{%
x  y
0 0.750423211451084
0.032258064516129 0.767942107242222
0.0645161290322581 0.780624775187343
0.0967741935483871 0.798977186083204
0.129032258064516 0.818822776046389
0.161290322580645 0.840617517901329
0.193548387096774 0.851414043898716
0.225806451612903 0.871933925341261
0.258064516129032 0.893646785867441
0.290322580645161 0.909977918602643
0.32258064516129 0.924222130594538
0.354838709677419 0.946145300885119
0.387096774193548 0.93864650940934
0.419354838709677 0.934140520288376
0.451612903225806 0.915153993216917
0.483870967741935 0.901877151918762
0.516129032258065 0.891208823531089
0.548387096774194 0.867999716353537
0.580645161290323 0.866516135966116
0.612903225806452 0.842726995182289
0.645161290322581 0.833044339161729
0.67741935483871 0.822484842265926
0.709677419354839 0.81254815963509
0.741935483870968 0.797615435501026
0.774193548387097 0.788805355755352
0.806451612903226 0.782893692577843
0.838709677419355 0.771447350115226
0.870967741935484 0.776028476914662
0.903225806451613 0.771511330369324
0.935483870967742 0.758009357782139
0.967741935483871 0.758535198110773
1 0.747840993338824
};
\addlegendentry{Weighted PCA}
\addplot [draw=red, fill=red, mark=square*, only marks, mark size=1.5pt]
table{%
x  y
0 0.750446084119576
0.032258064516129 0.751599832689736
0.0645161290322581 0.745582910483144
0.0967741935483871 0.749157086388162
0.129032258064516 0.748247866947687
0.161290322580645 0.749388607934651
0.193548387096774 0.751408633926084
0.225806451612903 0.750057404147824
0.258064516129032 0.747652790784086
0.290322580645161 0.746490656524183
0.32258064516129 0.752019994367123
0.354838709677419 0.743506120607932
0.387096774193548 0.753925252563213
0.419354838709677 0.753657362800476
0.451612903225806 0.751265179044738
0.483870967741935 0.753011729789267
0.516129032258065 0.754071260491485
0.548387096774194 0.75007547329019
0.580645161290323 0.755383154937598
0.612903225806452 0.752174790465882
0.645161290322581 0.750851814875378
0.67741935483871 0.751262224961117
0.709677419354839 0.755572752841667
0.741935483870968 0.747988228943163
0.774193548387097 0.749315131646505
0.806451612903226 0.749995695679842
0.838709677419355 0.744909678430838
0.870967741935484 0.753676668262882
0.903225806451613 0.752535083027454
0.935483870967742 0.746435660952136
0.967741935483871 0.753784313881318
1 0.747841022393646
};
\addlegendentry{Gradient Descent}
\addplot [draw=color0, fill=color0, mark=x, only marks]
table{%
x  y
0 0.621921350713139
0.032258064516129 0.635803428063128
0.0645161290322581 0.626717112485078
0.0967741935483871 0.61712297897985
0.129032258064516 0.613514179689596
0.161290322580645 0.612682841044922
0.193548387096774 0.62078263261961
0.225806451612903 0.617297527198462
0.258064516129032 0.613906061877634
0.290322580645161 0.608258803364184
0.32258064516129 0.622479367259711
0.354838709677419 0.612426609196666
0.387096774193548 0.628116462802927
0.419354838709677 0.62250659664105
0.451612903225806 0.620210625964245
0.483870967741935 0.619512429712917
0.516129032258065 0.619728276142777
0.548387096774194 0.617072983512675
0.580645161290323 0.628905269707922
0.612903225806452 0.618513769579149
0.645161290322581 0.617241247266547
0.67741935483871 0.615210495735869
0.709677419354839 0.623702223330041
0.741935483870968 0.617770399993836
0.774193548387097 0.618468380773733
0.806451612903226 0.622867945099137
0.838709677419355 0.613076362941792
0.870967741935484 0.623438218831568
0.903225806451613 0.626771556876521
0.935483870967742 0.611198491966033
0.967741935483871 0.637793483115843
1 0.615424731836837
};
\addlegendentry{AMP}
\path [draw=black, semithick]
(axis cs:0,0.617522635528674)
--(axis cs:1,0.617522635528674);

\end{axis}

\end{tikzpicture}

%% file: figures/fig3a.tex
\begin{tikzpicture}

\definecolor{color0}{rgb}{0.12156862745098,0.466666666666667,0.705882352941177}
\definecolor{color1}{rgb}{1,0.498039215686275,0.0549019607843137}
\definecolor{color2}{rgb}{0.172549019607843,0.627450980392157,0.172549019607843}
\definecolor{color3}{rgb}{0.83921568627451,0.152941176470588,0.156862745098039}
\definecolor{color4}{rgb}{0.580392156862745,0.403921568627451,0.741176470588235}

\begin{axis}[
legend cell align={left},
legend style={fill opacity=0.8, draw opacity=1, text opacity=1,
nodes={scale=0.8}, draw=white!80!black},
tick align=outside,
tick pos=left,
x grid style={white!69.0196078431373!black},
xlabel={\(\displaystyle \lambda\)},
xmin=-0.25, xmax=5.25,
xtick style={color=black},
y grid style={white!69.0196078431373!black},
ylabel={Diagonal block MSE},
title={Rademacher prior ($\bu$ and $\bv$)},
ymin=0.0355245032659438, ymax=1.04592740460638,
ytick style={color=black},
width=0.95\linewidth,
height=0.8\textwidth
]
\addplot [draw=color0, fill=color0, mark=x, only marks]
table{%
x  y
0.1 0.999999987614132
0.258064516129032 0.999999955031283
0.416129032258065 0.999983382384753
0.574193548387097 0.999989402720586
0.732258064516129 0.99814747207031
0.890322580645161 0.99480378985363
1.04838709677419 0.989293446289813
1.20645161290323 0.969366317509272
1.36451612903226 0.914352593106794
1.52258064516129 0.867481531198505
1.68064516129032 0.759690737708032
1.83870967741935 0.694068014782716
1.99677419354839 0.623720450642262
2.15483870967742 0.547209503831206
2.31290322580645 0.499295214701524
2.47096774193548 0.457677204977296
2.62903225806452 0.405516724927388
2.78709677419355 0.377954159277637
2.94516129032258 0.328440205646704
3.10322580645161 0.301532856557286
3.26129032258065 0.251190973087204
3.41935483870968 0.225302760876928
3.57741935483871 0.220077609061827
3.73548387096774 0.177988185240505
3.89354838709677 0.169197683589859
4.05161290322581 0.162973848173605
4.20967741935484 0.147979148173709
4.36774193548387 0.119000082912948
4.5258064516129 0.11566304523738
4.68387096774194 0.0957751291999771
4.84193548387097 0.088389886060264
5 0.089847834954261
};
\addlegendentry{$\alpha$ = 0.0}
\addplot [draw=color1, fill=color1, mark=x, only marks]
table{%
x  y
0.1 0.999999992173956
0.258064516129032 0.999999956541753
0.416129032258065 0.999998908318063
0.574193548387097 0.999995109376385
0.732258064516129 0.999507155883463
0.890322580645161 0.997385113939251
1.04838709677419 0.991611757242314
1.20645161290323 0.96565614326524
1.36451612903226 0.885923852426641
1.52258064516129 0.829633372382321
1.68064516129032 0.758670232039427
1.83870967741935 0.679810826879364
1.99677419354839 0.619529521239937
2.15483870967742 0.555636402310721
2.31290322580645 0.503566440847308
2.47096774193548 0.462439805731725
2.62903225806452 0.40405890195762
2.78709677419355 0.345274145280361
2.94516129032258 0.317534778403107
3.10322580645161 0.283532489329571
3.26129032258065 0.263119013619547
3.41935483870968 0.228010945321474
3.57741935483871 0.216720032017154
3.73548387096774 0.189274480599824
3.89354838709677 0.167540220051341
4.05161290322581 0.147954807417777
4.20967741935484 0.144735922328528
4.36774193548387 0.11707877195618
4.5258064516129 0.115810486226715
4.68387096774194 0.0980007929396549
4.84193548387097 0.0970928135824906
5 0.0862805259487824
};
\addlegendentry{$\alpha$ = 0.25}
\addplot [draw=color2, fill=color2, mark=x, only marks]
table{%
x  y
0.1 0.999999994571184
0.258064516129032 0.999999962846271
0.416129032258065 0.99999987376272
0.574193548387097 0.99999980922539
0.732258064516129 0.999776165985023
0.890322580645161 0.997632066989274
1.04838709677419 0.993421343760523
1.20645161290323 0.961561479108289
1.36451612903226 0.906445113526175
1.52258064516129 0.846763716948301
1.68064516129032 0.773567161787623
1.83870967741935 0.693577037527884
1.99677419354839 0.619103943728696
2.15483870967742 0.546300934733133
2.31290322580645 0.491659576084723
2.47096774193548 0.442777382193967
2.62903225806452 0.409598821349309
2.78709677419355 0.359006959016979
2.94516129032258 0.325979905864863
3.10322580645161 0.285758591924916
3.26129032258065 0.248477023721403
3.41935483870968 0.240508577751017
3.57741935483871 0.220195287058858
3.73548387096774 0.189943194272462
3.89354838709677 0.173775926231866
4.05161290322581 0.155013277361695
4.20967741935484 0.134519803289991
4.36774193548387 0.123632052332772
4.5258064516129 0.119333887108609
4.68387096774194 0.10196389963872
4.84193548387097 0.0952336909974897
5 0.0814519078723275
};
\addlegendentry{$\alpha$ = 0.5}
\addplot [draw=color3, fill=color3, mark=x, only marks]
table{%
x  y
0.1 0.999999995768896
0.258064516129032 0.999999959595073
0.416129032258065 0.999999883884146
0.574193548387097 0.999942621001463
0.732258064516129 0.999581910163743
0.890322580645161 0.999400095124966
1.04838709677419 0.990623716399278
1.20645161290323 0.968631837512819
1.36451612903226 0.906885890443359
1.52258064516129 0.846402342865793
1.68064516129032 0.762187192789167
1.83870967741935 0.679896015788611
1.99677419354839 0.629006836224456
2.15483870967742 0.568351128475841
2.31290322580645 0.495337928820815
2.47096774193548 0.435351351145315
2.62903225806452 0.39899871507733
2.78709677419355 0.362900053088779
2.94516129032258 0.313775055222758
3.10322580645161 0.302373704246787
3.26129032258065 0.253103714828573
3.41935483870968 0.235049222036223
3.57741935483871 0.205421007474178
3.73548387096774 0.184059710526711
3.89354838709677 0.170917495504976
4.05161290322581 0.158994025464093
4.20967741935484 0.141405301258679
4.36774193548387 0.127666885644878
4.5258064516129 0.121512120458765
4.68387096774194 0.101534663676813
4.84193548387097 0.0923069259476899
5 0.0847903008762978
};
\addlegendentry{$\alpha$ = 0.75}
\addplot [draw=color4, fill=color4, mark=x, only marks]
table{%
x  y
0.1 0.999999988892691
0.258064516129032 0.999999947705112
0.416129032258065 0.999999622874741
0.574193548387097 0.999971850481721
0.732258064516129 0.999983790187512
0.890322580645161 0.999527523778769
1.04838709677419 0.987538438058346
1.20645161290323 0.969764810408785
1.36451612903226 0.901660434671478
1.52258064516129 0.858295663179662
1.68064516129032 0.75660824833382
1.83870967741935 0.696266572364685
1.99677419354839 0.63597913575236
2.15483870967742 0.54040961568948
2.31290322580645 0.50579056288276
2.47096774193548 0.453037464859683
2.62903225806452 0.395992050491844
2.78709677419355 0.344405058036105
2.94516129032258 0.330227238686828
3.10322580645161 0.279519371164165
3.26129032258065 0.254402515521457
3.41935483870968 0.245093547416102
3.57741935483871 0.222730435297559
3.73548387096774 0.188950461001905
3.89354838709677 0.174124448098364
4.05161290322581 0.148919881057567
4.20967741935484 0.141598723735197
4.36774193548387 0.127712978489114
4.5258064516129 0.108987655926286
4.68387096774194 0.108806892309872
4.84193548387097 0.0912083511735144
5 0.0841154356004056
};
\addlegendentry{$\alpha$ = 1.0}
\addplot [semithick, color0]
table {%
0 1
0.00978473581213307 0.999999999999955
0.0195694716242661 0.999999999999954
0.0293542074363992 0.999999999999953
0.0391389432485323 0.999999999999952
0.0489236790606654 0.999999999999951
0.0587084148727984 0.99999999999995
0.0684931506849315 0.999999999999949
0.0782778864970646 0.999999999999948
0.0880626223091976 0.999999999999947
0.0978473581213307 0.999999999999946
0.107632093933464 0.999999999999945
0.117416829745597 0.999999999999943
0.12720156555773 0.999999999999942
0.136986301369863 0.999999999999941
0.146771037181996 0.999999999999939
0.156555772994129 0.999999999999938
0.166340508806262 0.999999999999936
0.176125244618395 0.999999999999935
0.185909980430528 0.999999999999933
0.195694716242661 0.999999999999932
0.205479452054795 0.99999999999993
0.215264187866928 0.999999999999928
0.225048923679061 0.999999999999927
0.234833659491194 0.999999999999925
0.244618395303327 0.999999999999923
0.25440313111546 0.999999999999921
0.264187866927593 0.999999999999919
0.273972602739726 0.999999999999916
0.283757338551859 0.999999999999914
0.293542074363992 0.999999999999912
0.303326810176125 0.999999999999909
0.313111545988258 0.999999999999907
0.322896281800391 0.999999999999904
0.332681017612524 0.999999999999901
0.342465753424658 0.999999999999898
0.352250489236791 0.999999999999895
0.362035225048924 0.999999999999892
0.371819960861057 0.999999999999888
0.38160469667319 0.999999999999885
0.391389432485323 0.999999999999881
0.401174168297456 0.999999999999877
0.410958904109589 0.999999999999873
0.420743639921722 0.999999999999869
0.430528375733855 0.999999999999864
0.440313111545988 0.999999999999859
0.450097847358121 0.999999999999854
0.459882583170254 0.999999999999849
0.469667318982387 0.999999999999843
0.479452054794521 0.999999999999837
0.489236790606654 0.999999999999831
0.499021526418787 0.999999999999824
0.50880626223092 0.999999999999817
0.518590998043053 0.99999999999981
0.528375733855186 0.999999999999802
0.538160469667319 0.999999999999793
0.547945205479452 0.999999999999784
0.557729941291585 0.999999999999775
0.567514677103718 0.999999999999764
0.577299412915851 0.999999999999753
0.587084148727984 0.999999999999741
0.596868884540117 0.999999999999729
0.60665362035225 0.999999999999715
0.616438356164384 0.9999999999997
0.626223091976517 0.999999999999684
0.63600782778865 0.999999999999667
0.645792563600783 0.999999999999648
0.655577299412916 0.999999999999628
0.665362035225049 0.999999999999606
0.675146771037182 0.999999999999582
0.684931506849315 0.999999999999556
0.694716242661448 0.999999999999527
0.704500978473581 0.999999999999495
0.714285714285714 0.99999999999946
0.724070450097847 0.999999999999421
0.73385518590998 0.999999999999377
0.743639921722113 0.999999999999329
0.753424657534246 0.999999999999275
0.76320939334638 0.999999999999213
0.772994129158513 0.999999999999144
0.782778864970646 0.999999999999065
0.792563600782779 0.999999999998975
0.802348336594912 0.999999999998871
0.812133072407045 0.99999999999875
0.821917808219178 0.999999999998609
0.831702544031311 0.999999999998443
0.841487279843444 0.999999999998245
0.851272015655577 0.999999999998006
0.86105675146771 0.999999999997715
0.870841487279843 0.999999999997356
0.880626223091976 0.999999999996905
0.89041095890411 0.999999999996327
0.900195694716243 0.999999999995572
0.909980430528376 0.999999999994557
0.919765166340509 0.999999999993149
0.929549902152642 0.999999999991114
0.939334637964775 0.999999999988016
0.949119373776908 0.999999999982964
0.958904109589041 0.999999999973888
0.968688845401174 0.999999999955025
0.978473581213307 0.999999999904888
0.98825831702544 0.999999999680962
0.998043052837573 0.999999989580714
1.00782778864971 0.999938827634015
1.01761252446184 0.999693378704751
1.02739726027397 0.999263627329668
1.03718199608611 0.998653921113148
1.04696673189824 0.997868684822219
1.05675146771037 0.996912521389538
1.0665362035225 0.995790089590106
1.07632093933464 0.994506148608361
1.08610567514677 0.99306532044499
1.0958904109589 0.991472477875989
1.10567514677104 0.989732368006899
1.11545988258317 0.987849725761528
1.1252446183953 0.985829256027271
1.13502935420744 0.98367567687269
1.14481409001957 0.98139355521045
1.1545988258317 0.978987456062306
1.16438356164384 0.976461869896056
1.17416829745597 0.973821234059869
1.1839530332681 0.971069860365427
1.19373776908023 0.968212027792243
1.20352250489237 0.965251937554238
1.2133072407045 0.962193639010875
1.22309197651663 0.959041153282885
1.23287671232877 0.955798396618022
1.2426614481409 0.95246919231245
1.25244618395303 0.949057283685751
1.26223091976517 0.945566298896452
1.2720156555773 0.941999792652151
1.28180039138943 0.938361200638849
1.29158512720157 0.934653922695623
1.3013698630137 0.930881196187463
1.31115459882583 0.927046224701463
1.32093933463796 0.923152104686192
1.3307240704501 0.919201844906605
1.34050880626223 0.915198367645213
1.35029354207436 0.911144509999182
1.3600782778865 0.907043025258833
1.36986301369863 0.902896584354536
1.37964774951076 0.898707797225806
1.3894324853229 0.894479135765508
1.39921722113503 0.890213052058501
1.40900195694716 0.885911903050127
1.4187866927593 0.881577971218998
1.42857142857143 0.877213476811587
1.43835616438356 0.872820547590003
1.44814090019569 0.86840124316845
1.45792563600783 0.863957574325384
1.46771037181996 0.859491476076492
1.47749510763209 0.855004818913032
1.48727984344423 0.850499431110174
1.49706457925636 0.845977018463297
1.50684931506849 0.841439285813142
1.51663405088063 0.836887865107064
1.52641878669276 0.832324326764411
1.53620352250489 0.827750194444138
1.54598825831703 0.82316692506587
1.55577299412916 0.818575944010416
1.56555772994129 0.813978579922093
1.57534246575342 0.809376163708075
1.58512720156556 0.804769959718536
1.59491193737769 0.800161186001163
1.60469667318982 0.795551015649282
1.61448140900196 0.790940578120833
1.62426614481409 0.786330960527948
1.63405088062622 0.781723205444525
1.64383561643836 0.777118329727902
1.65362035225049 0.772517302776893
1.66340508806262 0.767921034717107
1.67318982387476 0.763330429547152
1.68297455968689 0.758746345246202
1.69275929549902 0.754169610121062
1.70254403131115 0.749601011791362
1.71232876712329 0.745041309830433
1.72211350293542 0.740491234523274
1.73189823874755 0.735951478763453
1.74168297455969 0.731422719189076
1.75146771037182 0.726905624994122
1.76125244618395 0.722400741406318
1.77103718199609 0.717908691091078
1.78082191780822 0.713430035527631
1.79060665362035 0.708965307016072
1.80039138943249 0.704515009245265
1.81017612524462 0.700079662571022
1.81996086105675 0.695659716936511
1.82974559686888 0.691255617851669
1.83953033268102 0.686867789857659
1.84931506849315 0.682496638591965
1.85909980430528 0.67814254600523
1.86888454011742 0.673805879136618
1.87866927592955 0.669486985368042
1.88845401174168 0.66518619437238
1.89823874755382 0.660903818708225
1.90802348336595 0.65664015439687
1.91780821917808 0.652395481481568
1.92759295499022 0.648170064570065
1.93737769080235 0.643964153360512
1.94716242661448 0.639777983151322
1.95694716242661 0.635611775335397
1.96673189823875 0.631465747775944
1.97651663405088 0.627340075972389
1.98630136986301 0.623234952031732
1.99608610567515 0.619150560570936
2.00587084148728 0.615087032351468
2.01565557729941 0.611044536526001
2.02544031311155 0.607023194380494
2.03522504892368 0.603023134253934
2.04500978473581 0.599044487057586
2.05479452054794 0.595087345773956
2.06457925636008 0.591151816241453
2.07436399217221 0.587237976925385
2.08414872798434 0.583345911744862
2.09393346379648 0.579475694900948
2.10371819960861 0.575627382498742
2.11350293542074 0.571801035631633
2.12328767123288 0.567996696071668
2.13307240704501 0.564214409188991
2.14285714285714 0.560454203558628
2.15264187866928 0.556716109800445
2.16242661448141 0.553000145956604
2.17221135029354 0.549306326338253
2.18199608610567 0.545634657264832
2.19178082191781 0.541985142411548
2.20156555772994 0.538357782677769
2.21135029354207 0.534752561095061
2.22113502935421 0.53116946806048
2.23091976516634 0.527608484958456
2.24070450097847 0.524069592428557
2.25048923679061 0.520552754852161
2.26027397260274 0.517057945205047
2.27005870841487 0.51358512459921
2.27984344422701 0.510134254019506
2.28962818003914 0.506705293186551
2.29941291585127 0.503298186741669
2.30919765166341 0.499912890477744
2.31898238747554 0.496549343165239
2.32876712328767 0.49320749019865
2.3385518590998 0.489887273854465
2.34833659491194 0.486588623910802
2.35812133072407 0.483311476815164
2.3679060665362 0.480055766154062
2.37769080234834 0.476821414139005
2.38747553816047 0.473608367348058
2.3972602739726 0.470416513347036
2.40704500978474 0.467245791029463
2.41682974559687 0.464096121007586
2.426614481409 0.460967415381122
2.43639921722114 0.457859592717894
2.44618395303327 0.45477256576894
2.4559686888454 0.451706245896739
2.46575342465753 0.448660542886088
2.47553816046967 0.445635364986764
2.4853228962818 0.442630618985426
2.49510763209393 0.43964621028212
2.50489236790607 0.43668204296626
2.5146771037182 0.433738019891062
2.52446183953033 0.4308140427452
2.53424657534247 0.427910012122666
2.5440313111546 0.425025827590016
2.55381604696673 0.422161387751372
2.56360078277886 0.419316590311165
2.573385518591 0.416491332134705
2.58317025440313 0.413685509306613
2.59295499021526 0.410899017187205
2.6027397260274 0.408131750466883
2.61252446183953 0.405383603218583
2.62230919765166 0.402654468948365
2.6320939334638 0.399944240644185
2.64187866927593 0.39725281082291
2.65166340508806 0.394580071575646
2.6614481409002 0.391925914611411
2.67123287671233 0.389290237845363
2.68101761252446 0.386672919397561
2.69080234833659 0.384073856482578
2.70058708414873 0.381492939686835
2.71037181996086 0.378930059410183
2.72015655577299 0.376385105900763
2.72994129158513 0.373857969288564
2.73972602739726 0.371348539617711
2.74951076320939 0.368856706877533
2.75929549902153 0.366382361032439
2.76908023483366 0.363925392050653
2.77886497064579 0.361485692595302
2.78864970645793 0.359063149943895
2.79843444227006 0.356657653115954
2.80821917808219 0.354269093065331
2.81800391389432 0.35189736180344
2.82778864970646 0.349542347089595
2.83757338551859 0.347203941315912
2.84735812133072 0.344882035389548
2.85714285714286 0.342576523304818
2.86692759295499 0.340287290737278
2.87671232876712 0.338014232114279
2.88649706457926 0.335757239320632
2.89628180039139 0.333516204529469
2.90606653620352 0.331291024040046
2.91585127201566 0.32908158308162
2.92563600782779 0.326887778533523
2.93542074363992 0.324709503876129
2.94520547945205 0.322546652951787
2.95499021526419 0.320399119977638
2.96477495107632 0.318266802707981
2.97455968688845 0.316149589906427
2.98434442270059 0.314047380075641
2.99412915851272 0.311960071792417
3.00391389432485 0.309887555885302
3.01369863013699 0.307829731749517
3.02348336594912 0.305786496537488
3.03326810176125 0.303757747859303
3.04305283757339 0.301743386478334
3.05283757338552 0.299743305620097
3.06262230919765 0.29775740695223
3.07240704500978 0.295785589992871
3.08219178082192 0.293827757226225
3.09197651663405 0.291883804267966
3.10176125244618 0.289953634563289
3.11154598825832 0.288037149642904
3.12133072407045 0.286134253736749
3.13111545988258 0.284244845089152
3.14090019569472 0.282368828937501
3.15068493150685 0.280506110698196
3.16046966731898 0.278656590979821
3.17025440313112 0.27682017623749
3.18003913894325 0.274996773042503
3.18982387475538 0.273186291426899
3.19960861056751 0.271388625360419
3.20939334637965 0.26960368849229
3.21917808219178 0.267831388249605
3.22896281800391 0.266071632612845
3.23874755381605 0.264324330117497
3.24853228962818 0.262589389855765
3.25831702544031 0.260866721477998
3.26810176125245 0.259156235193714
3.27788649706458 0.257457841843882
3.28767123287671 0.255771452620479
3.29745596868885 0.254096980417963
3.30724070450098 0.252434335841207
3.31702544031311 0.250783432824987
3.32681017612524 0.249144184949919
3.33659491193738 0.247516506359243
3.34637964774951 0.245900311758887
3.35616438356164 0.244295516417142
3.36594911937378 0.242702036164257
3.37573385518591 0.241119787391895
3.38551859099804 0.23954869042513
3.39530332681018 0.237988656089583
3.40508806262231 0.236439605771656
3.41487279843444 0.234901459656195
3.42465753424658 0.233374133842843
3.43444227005871 0.231857549607337
3.44422700587084 0.230351627244706
3.45401174168297 0.228856287604642
3.46379647749511 0.22737145208837
3.47358121330724 0.225897042647072
3.48336594911937 0.22443298178023
3.49315068493151 0.222979192533881
3.50293542074364 0.221535598498817
3.51272015655577 0.220102123808691
3.52250489236791 0.218678693138075
3.53228962818004 0.217265231700437
3.54207436399217 0.215861669895088
3.55185909980431 0.214467926295759
3.56164383561644 0.213083929300659
3.57142857142857 0.211709609005661
3.5812133072407 0.210344890777047
3.59099804305284 0.208989704260573
3.60078277886497 0.207643978380731
3.6105675146771 0.20630764417026
3.62035225048924 0.204980628443908
3.63013698630137 0.203662864075224
3.6399217221135 0.202354281613901
3.64970645792564 0.201054810985176
3.65949119373777 0.199764386369758
3.6692759295499 0.198482938339077
3.67906066536204 0.197210400877342
3.68884540117417 0.195946707503893
3.6986301369863 0.194691794357551
3.70841487279843 0.193445591739271
3.71819960861057 0.192208036740281
3.7279843444227 0.190979064852349
3.73776908023483 0.189758612053358
3.74755381604697 0.188546614804472
3.7573385518591 0.18734301004728
3.76712328767123 0.186147735200936
3.77690802348337 0.184960728159277
3.7866927592955 0.183781933044668
3.79647749510763 0.182611277269486
3.80626223091976 0.181448705800853
3.8160469667319 0.180294158402345
3.82583170254403 0.17914757529767
3.83561643835616 0.178008897865961
3.8454011741683 0.176878065856949
3.85518590998043 0.175755021544464
3.86497064579256 0.174639706963352
3.8747553816047 0.173532064593874
3.88454011741683 0.172432037358756
3.89432485322896 0.171339568620234
3.9041095890411 0.170254602177099
3.91389432485323 0.16917708226175
3.92367906066536 0.168106953537238
3.93346379647749 0.16704416109432
3.94324853228963 0.165988650450409
3.95303326810176 0.16494037009856
3.96281800391389 0.163899260619052
3.97260273972603 0.162865272687401
3.98238747553816 0.161838352804003
3.99217221135029 0.160818448564119
4.00195694716243 0.159805507970188
4.01174168297456 0.158799479428918
4.02152641878669 0.157800311748383
4.03131115459883 0.15680795413513
4.04109589041096 0.155822356191294
4.05088062622309 0.154843473869571
4.06066536203523 0.15387124573149
4.07045009784736 0.152905628415259
4.08023483365949 0.151946573076708
4.09001956947162 0.150994031252972
4.09980430528376 0.150047954216795
4.10958904109589 0.149108295533114
4.11937377690802 0.148175007235015
4.12915851272016 0.147248048493302
4.13894324853229 0.146327360530169
4.14872798434442 0.145412903155539
4.15851272015656 0.144504630497287
4.16829745596869 0.143602497042148
4.17808219178082 0.142706457633228
4.18786692759295 0.141816467467205
4.19765166340509 0.140932482091635
4.20743639921722 0.140054457402328
4.21722113502935 0.139182349640321
4.22700587084149 0.138316116610026
4.23679060665362 0.137455712813375
4.24657534246575 0.136601097387307
4.25636007827789 0.1357522265932
4.26614481409002 0.134909060127536
4.27592954990215 0.134071554056628
4.28571428571429 0.133239668055059
4.29549902152642 0.132413361033708
4.30528375733855 0.131592593704969
4.31506849315068 0.13077732268346
4.32485322896282 0.129967509297161
4.33463796477495 0.129163113730101
4.34442270058708 0.128364096478376
4.35420743639922 0.12757042011905
4.36399217221135 0.126782042247564
4.37377690802348 0.125998926027835
4.38356164383562 0.125221033179729
4.39334637964775 0.124448325723003
4.40313111545988 0.123680765974918
4.41291585127202 0.122918318538278
4.42270058708415 0.122160942365599
4.43248532289628 0.12140860261506
4.44227005870841 0.120661262770376
4.45205479452055 0.119918886601062
4.46183953033268 0.119181438160136
4.47162426614481 0.11844888389379
4.48140900195695 0.117721184220975
4.49119373776908 0.116998306114308
4.50097847358121 0.116280214738223
4.51076320939335 0.115566893878093
4.52054794520548 0.114858254194729
4.53033268101761 0.114154316708796
4.54011741682975 0.113455048040171
4.54990215264188 0.112760377501866
4.55968688845401 0.112070290325391
4.56947162426614 0.11138475353554
4.57925636007828 0.110703734414326
4.58904109589041 0.110027200498893
4.59882583170254 0.109355119579441
4.60861056751468 0.108687459697171
4.61839530332681 0.108024189142237
4.62818003913894 0.107365276451729
4.63796477495108 0.106710690407651
4.64774951076321 0.10606040003493
4.65753424657534 0.105414374599435
4.66731898238748 0.104772583606008
4.67710371819961 0.104134996796517
4.68688845401174 0.103501584147917
4.69667318982387 0.10287231587033
4.70645792563601 0.102247162405138
4.71624266144814 0.101626094423095
4.72602739726027 0.101009082822447
4.73581213307241 0.100396098727071
4.74559686888454 0.0997871134846307
4.75538160469667 0.0991820986647379
4.76516634050881 0.09858102605714
4.77495107632094 0.0979838676699152
4.78473581213307 0.0973905957276801
4.79452054794521 0.0968011826698169
4.80430528375734 0.0962156011487125
4.81409001956947 0.0956338240280115
4.8238747553816 0.095055824380882
4.83365949119374 0.0944815754882964
4.84344422700587 0.0939110508373283
4.853228962818 0.09334422411946
4.86301369863014 0.0927810692289049
4.87279843444227 0.0922215602609429
4.8825831702544 0.0916656715102699
4.89236790606654 0.0911133774693607
4.90215264187867 0.0905646528268469
4.9119373776908 0.0900194581605177
4.92172211350293 0.0894777969662771
4.93150684931507 0.0889396303950239
4.9412915851272 0.0884049339040154
4.95107632093933 0.0878736831387411
4.96086105675147 0.087345855976751
4.9706457925636 0.0868214243713227
4.98043052837573 0.0863003665424059
4.99021526418787 0.0857826588731002
5 0.0852682779271359
};
\addlegendentry{MMSE}
\end{axis}

\end{tikzpicture}